\documentclass[10pt,twocolumn,letterpaper]{article}

\usepackage[pagenumbers]{cvpr} % To force page numbers, e.g. for an arXiv version

\usepackage{graphicx}
\usepackage{amsmath}
\usepackage{amssymb}
\usepackage{booktabs}

\usepackage{appendix}

\usepackage{times}
\usepackage{epsfig}
\usepackage{mathtools}
\usepackage[dvipsnames]{xcolor}

\usepackage{kotex}
\usepackage{multirow}
\usepackage{ftnxtra}
\usepackage{fnpos}
\usepackage{comment}
\usepackage{adjustbox}

\usepackage{algorithm}
\usepackage{algorithmic}

\usepackage[pagebackref,breaklinks,colorlinks]{hyperref}

\usepackage[capitalize]{cleveref}
\crefname{section}{Sec.}{Secs.}
\Crefname{section}{Section}{Sections}
\Crefname{table}{Table}{Tables}
\crefname{table}{Tab.}{Tabs.}

\def\cvprPaperID{7796} % *** Enter the CVPR Paper ID here
\def\confName{CVPR}
\def\confYear{2022}

\begin{document}

\title{Imposing Consistency for Optical Flow Estimation}
% \title{ Building on Consistency for Optical Flow Estimation}
%\title{ Consistency Learning for Optical Flow Estimation}
% \title{Learning Consistency Enhancements for Optical Flow Estimation}

% Fatih Porikli
\author{
Jisoo Jeong$^1$~~~
Jamie Menjay Lin$^{2, \dagger}$~~~
% Jamie Menjay Lin$^{2, \dagger}$~~~
Fatih Porikli$^{1}$~~~
Nojun Kwak$^{3, \ddagger}$~~~
\smallskip
\\
$^1$Qualcomm AI Research$^{*}$~~~
% \\{\tt\small\{jisojeon,fporikli\}@qti.qualcomm.com} 
% \and
$^2$Google Research~~~
% \\{\tt\small jmlin@google.com}
% \and
$^3$Seoul National University~~~
% \\{\tt\small nojunk@snu.ac.kr}
\smallskip
\\
{\tt\small\{jisojeon,fporikli\}@qti.qualcomm.com}~~~
{\tt\small jmlin@google.com}~~~
{\tt\small nojunk@snu.ac.kr}
}

% \author{
% Jisoo Jeong\\
% {\normalsize Qualcomm AI Research}\\

% % {\tt\small jisojeon@qti.qualcomm.com}
% % Qualcomm AI Research\\
% % {\tt\small jisojeon@qti.qualcomm.com}
% \and
% Jamie Menjay Lin$^{*}$\\
% {\normalsize Google Research}\\
% % Google Research \\
% % {\tt\small jmlin@google.com}
% \and
% Fatih Porikli\\
% {\normalsize Qualcomm AI Research}\\
% % Qualcomm AI Research\\
% % {\tt\small fporikli@qti.qualcomm.com}
% \and
% Nojun Kwak\\
% {\normalsize Seoul National University}\\
% % Seoul National University\\
% % {\tt\small nojunk@snu.ac.kr}
% }

% {\tt\small nojunk@snu.ac.kr}
% {\tt\small fporikli@qti.qualcomm.com}
% {\tt\small jmlin@google.com}
% {\tt\small jisojeon@qti.qualcomm.com}
% Google Research \\
% Seoul National University\\
% Qualcomm AI Research\\

% \and
\maketitle

%%%%%%%%% ABSTRACT
\begin{abstract}

% \jl{Please check spelling of Fatih's last name :)}

Imposing consistency through proxy tasks has been shown to enhance data-driven learning and enable self-supervision in various tasks. This paper introduces novel and effective consistency strategies for optical flow estimation, a problem where labels from real-world data are very challenging to derive. More specifically, we propose occlusion consistency and zero forcing in the forms of self-supervised learning and transformation consistency in the form of semi-supervised learning. We apply these consistency techniques in a way that the network model learns to describe pixel-level motions better while requiring no additional annotations. We demonstrate that our consistency strategies applied to a strong baseline network model using the original datasets and labels provide further improvements, attaining the state-of-the-art results on the KITTI-2015 scene flow benchmark in the non-stereo category. Our method achieves the best foreground accuracy ($4.33\%$ in Fl-all) over both the stereo and non-stereo categories, even though using only monocular image inputs.

\end{abstract}

%%%%%%%%% Sec. 1
\section{Introduction}
\label{sec:intro}

{\let\thefootnote\relax\footnotetext{{
\hspace{-6.5mm} * Qualcomm AI Research is an initiative of Qualcomm Technologies, Inc.\\
\hspace{-6.5mm} $\dagger$ This work was done while at Qualcomm AI Research. \\
$\ddagger$ Nojun Kwak was supported by the National Research
Foundation of Korea (NRF) grant funded by the Korea government (2021R1A2C3006659).}}}

Optical flow characterizes dense displacements between corresponding pixels across images, e.g. between two consecutive frames in a video \cite{dosovitskiy2015flownet, ilg2017flownet, sun2019models, teed2020raft}. It is widely employed in video analysis applications including video compression~\cite{wu2018video, lu2019dvc}, action recognition~\cite{lee2018motion, cai2019temporal}, video denoising~\cite{bodduna2021removing, dewil2021self}, and object tracking~\cite{kale2015moving, zhou2018deeptam}, to point out a few.

As important as its, optical flow estimation comes with significant challenges. Occlusions due to camera and object motions present one inherent difficulty, where a part of the scene is visible in one but not in the other image of the pair. Several methods addressed this problem by explicitly estimating regions to be excluded \cite{meister2017unflow, zhao2020maskflownet}, by applying self-supervision \cite{liu2019selflow}, or by incorporating contextual information \cite{teed2020raft}. These methods, however, had limited reception since they rely on multiple forward-backward iterations for predicting occlusion areas \cite{sundaram2010dense, meister2017unflow} or fail for larger occlusions.

\begin{figure}[t]
\begin{center}
%\begin{tabular}{c}
\includegraphics[width=0.95\linewidth]{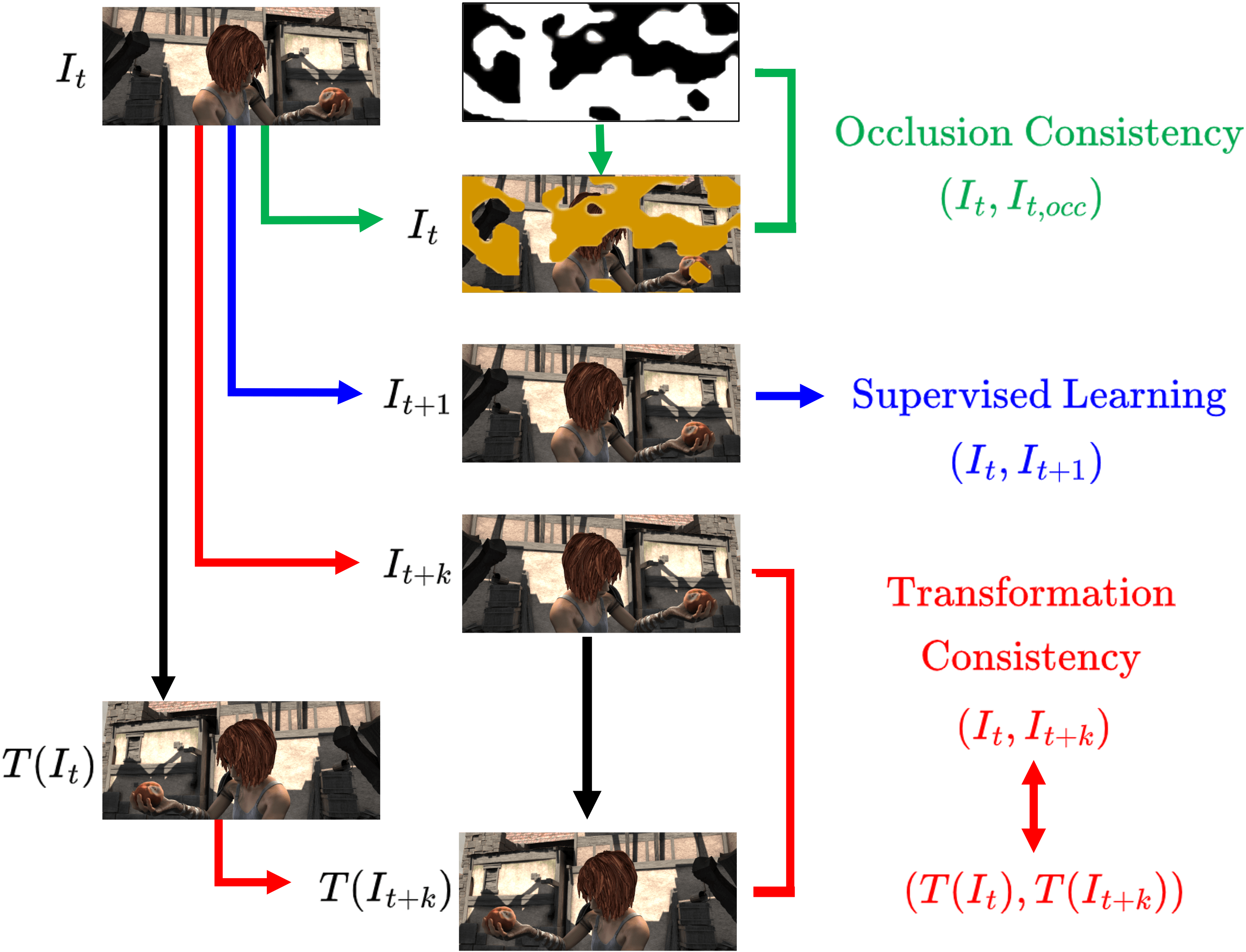}\\
% \includegraphics[width=8cm]{image/intro_dataset5.png}\\
%\end{tabular}
\vspace{-0.7mm}
\end{center}
\caption{During training, we enforce occlusion consistency with self-supervision by applying random occlusion patterns and imposing the network to detect the regions under occlusion between consecutive images ($I_{t},I_{t+1}$). We also employ transformation consistency (equivariance to geometric transformations) in a semi-supervised manner for an image pair ($I_{t}, I_{t+k}$) and the transformed pair ($T(I_{t}), T(I_{t+k})$) with $k \ge 1$. 
% \fp{update the figure with the new notation!!} 
}
\label{fig:our_scheme}
\vspace{-3mm}
\end{figure}

Obtaining precise annotations for optical flow is another challenge that directly impacts the learning performance. Since pixel-level motion annotation requires specialized and costly data acquisition systems, and in many cases, such annotations do not support high precision and spatial resolution, optical flow datasets are limited in number, variety, and degree of realism \cite{dosovitskiy2015flownet, ilg2017flownet}. The need for large-scale real-world datasets, therefore, becomes a bottleneck.

To mitigate the annotation issues, unsupervised learning~\cite{meister2017unflow, wang2018occlusion, janai2018unsupervised, jonschkowski2020matters} and semi-supervised learning~\cite{lai2017semi, yan2020optical} methods have been proposed in the past. Unsupervised learning schemes, however, typically result in degraded performance, lagging behind fully supervised learning counterparts~\cite{wang2018occlusion, liu2019ddflow, jonschkowski2020matters}. In comparison, semi-supervised learning~\cite{lai2017semi} may offer potential performance gains with data augmentation along with generative adversarial networks~\cite{goodfellow2014generative}.

In this paper, we introduce two consistency strategies for optical flow estimation to address these challenges as depicted in Fig.~\ref{fig:our_scheme}. First, we propose occlusion consistency that generates a random occlusion mask, which is used to create additional image pairs, and constrains the network to predict the mask and a zero-forced flow field in a self-supervised manner. Unlike other approaches, our occlusion consistency allows generating occlusion ground truth without forward-backward iterations. Although this intuitive strategy is simple, it enables the network not to confuse occlusion patterns as motion indicators without losing its representative capacity for the unoccluded image regions. It also helps the network to derive more informative features for the partially occluded regions within local receptive fields of the kernels without requiring additional labeling.

We also incorporate a transformation-based consistency regularization that has been shown useful in semi-supervised image classification and object detection tasks~\cite{laine2016temporal, tarvainen2017mean, oliver2018realistic, jeong2019consistency, jeong2021interpolation}. This strategy helps the model impose \textit{equivariance} through such consistency regularization. We apply whole-image geometric transformations including flippings, translations, and rotations. 
Then we restore the transformation before evaluating the overall transformation consistency losses.
While our transformation consistency is derived with two passes of forward flow estimation, the cycle consistency \cite{wang2019learning} is computed with one pass of forward and the other pass of backward flow estimation.
To the best of our knowledge, this is the first attempt to impose equivariance through consistency regularization for optical flow estimation. Note that our approach is different from conventional data augmentation schemes, which expand training samples without imposition of sophisticated consistency losses during training.

Our proposed self- and semi-supervised consistency learning strategies not only complement the previous state-of-the-art RAFT \cite{teed2020raft} baseline, but enable significant improvement in the model accuracy performance as evidenced in our experiment results.
Our proposed method achieves the new state-of-the-art accuracies and has ranked at the top of the KITTI-2015 scene flow non-stereo leaderboard (Ours: {$4.33\%$, $6.01\%$, $3.99\%$} vs. RAFT: {$5.10\%$, $6.87\%$, $4.74\%$} in Fl-all, Fl-fg, and Fl-bg, respectively). Our training with consistency strategies can potentially be adapted to other dense prediction tasks.

In summary, our main contributions are as follows:
\vspace{-0.5mm}
\begin{itemize}
\vspace{-0.5mm}
\item We propose a novel occlusion consistency strategy, which 
facilitates learning occlusion-robust representations efficiently in a self-supervised manner. 

\vspace{-0.5mm}
\item We incorporate transformation consistency equivariance enabling learning from a more diverse set of image pairs without additional labeling.

\vspace{-0.5mm}
\item Applying these two consistency strategies jointly in training and integrating an occlusion estimation channel in the architecture, our model generates superior results over its baseline achieving state-of-the-art performance in the KITTI-2015 scene flow non-stereo monocular dataset.

\end{itemize}
%%%%%%%%% Sec. 1

%%%%%%%%% Sec. 2
\section{Related Work}% \fp{we should shorten this part!!!}}
\label{sec:related}

% \subsection{Optical Flow}
% \label{rel:optical}

\noindent \textbf{Optical Flow:}
%There are classical formulations of optical flow 
Classic solutions have been studied for decades \cite{horn1981determining, brox2004high}, and recent advancements have been made with deep learning methods \cite{dosovitskiy2015flownet, ilg2017flownet, ranjan2017optical, sun2018pwc, zhao2020maskflownet, teed2020raft}. 
RAFT~\cite{teed2020raft} demonstrates notable improvement by extracting per-pixel features from the corresponding image pair $(I_{t}, I_{t+1})$, building multi-scale 4-dimensional correlation volumes for all pixel pairs, and iteratively adjust the flow estimates through a refinement module with gated recurrent units (GRUs) \cite{cho2014learning} with repeated lookups in the correlation volume. 
The loss is computed between the ground truth optical flow $f(I_t, I_{t+1})$ and the predicted optical flow $\tilde{f^i}(I_t,I_{t+1})$ in each iteration $i$ with $\ell_1$ norm 
\begin{equation} 
\label{related:raft}
\mathcal{L}_{RAFT} = \sum_{i=1}^N \gamma^{N-i} \left\| f(I_t, I_{t+1}) - \tilde{f}^{i}(I_t, I_{t+1}) \right\| _{1},
\end{equation}
where $N$ is the number of GRU iterations and $\gamma$ is a decay factor ($\gamma$ $<$ 1). The final predicted flow is then $\tilde{f}(I_t, I_{t+1}) = \tilde{f}^{N}(I_t, I_{t+1})$, the prediction after all iterations.
\vspace{3mm}
\\
% \\
\textbf{Methods for Occlusion Handling:} 
UnFlow \cite{meister2017unflow} identifies occlusions with the forward-backward constraint assumption \cite{sundaram2010dense} and excludes the occlusion area during training. For the forward-backward constraint, a bidirectional optical flow is required, and the errors could accumulate and propagate, partially due to the discretization of continuous values in the estimates. Self-supervised learning has also been introduced in recent works for optical flow estimation. SelFlow~\cite{liu2019selflow}, as an example,  performs flow estimation for non-occluded regions and uses these predictions to estimate flows in occluded regions. However, it requires four optical flow inferences (forward/backward$\times$occlusion/non-occlusion pairs) and significantly increases computational and memory costs to obtain occlusion maps and non-occlusion/occlusion flows. Maskflownet \cite{zhao2020maskflownet} proposes a learnable occlusion mask, which is applied to the next image frame $I_{t+1}$ when calculating the correlation between the features of $I_{t}$ and $I_{t+1}$. Recent studies \cite{hur2019iterative, jonschkowski2020matters} also propose predicting the occlusion mask with an additional channel, and we adopt this approach.

Another solution is to integrate contextual information. Recently, RAFT~\cite{teed2020raft} presented a context sub-network to incorporate neighborhood pixels' information. By assuming the pixels in an object or segment to have a similar flow, it refines the estimated flow fields in occlusion areas. However, as shown in Fig.~\ref{exp:example_image} (RAFT results), the matched parts can be incorrectly updated in case of severe occlusions. We analyze contextual information in more detail in the following subsection.

In contrast to previous algorithms, our method generates occlusion itself and enforces the network to predict the occlusion areas without multiple inferences.\footnote{Note that our contribution is not simply adding a channel but proposing a new scheme that generates and trains occlusion without occlusion prediction.}
\begin{figure}[t]
\begin{center}
\begin{tabular}{c}
\includegraphics[width=8cm]{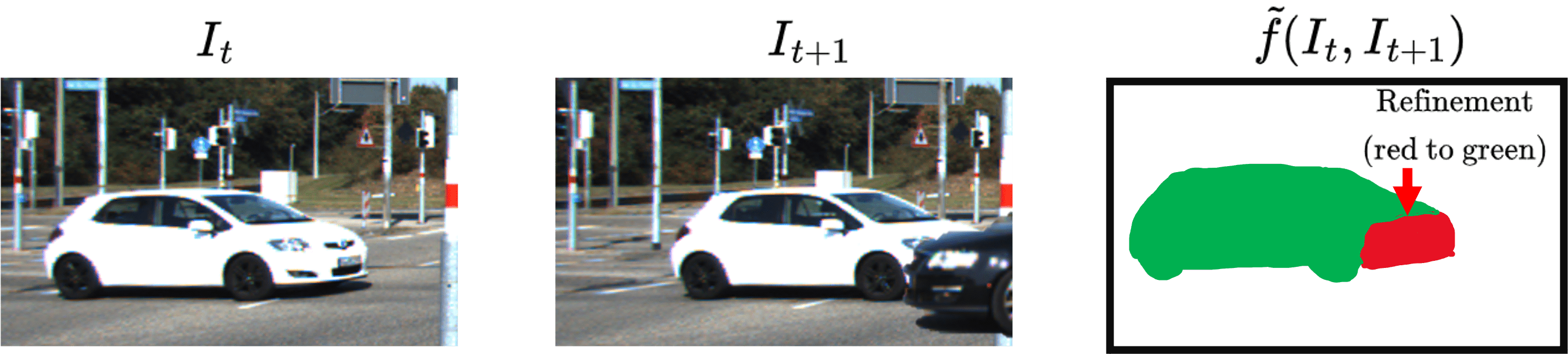} 
\\
(a) minor occlusion
\\
\includegraphics[width=8cm]{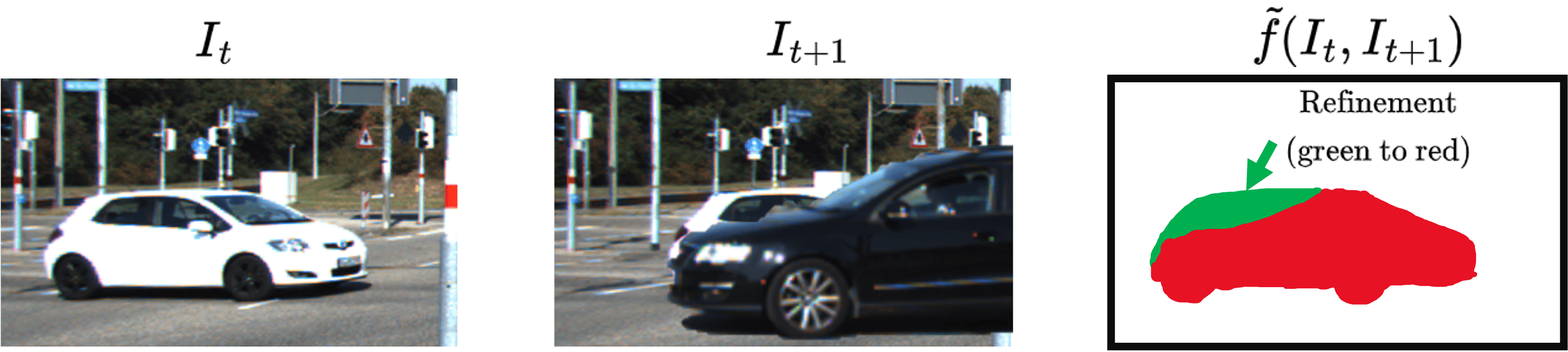}
\\
(b) major occlusion
% (a) & (b)
\end{tabular}
\vspace{-5mm}
\end{center}
\caption{Illustration of occlusion problems: (a) In a case of minor occlusion, incorrect optical flow estimations for the occlusion area can be corrected using larger spatial context (red to green). (b) However, in a case of major occlusion, the occlusion area can degenerate the accurately estimated optical flow of the smaller visible region (green to red) 
% \fp{update the figure with new notation!!}. 
}
% \vspace{-2mm}
\label{method:occlusion}
\end{figure}
\vspace{3mm}
\\
\noindent \textbf{Contextual Information: }Using context to regularize estimations within an image segment can improve optical flow as \cite{teed2020raft} intended with the context sub-network. However, such a regularization needs to be imposed while keeping the degree of occlusion in mind. Figure~\ref{method:occlusion} shows an example. In the case of minor occlusions, most pixels (in green) in a contextual segment (car) are likely to be estimated correctly. Here, the context sub-network may provide adequate support over the refinement iterations. On the other hand, in the case of major occlusions, the dominating portion of the occlusion region (in red) can be biased towards incorrect context, creating possibly significant deterioration in the correspondence estimation. RAFT estimation in Figure~\ref{exp:example_image} gives a real example of this problem occurring under a major occlusion. To tackle this problem, we propose the occlusion consistency strategy, as described in Section \ref{subsec:self}. 
\vspace{2mm}
\\
% \subsection{Self-Supervised Learning}
% \label{rel:self}
\textbf{Self-Supervised Learning:} By defining pretext tasks for unlabeled data and then using them to pretrain models, self-supervision allows making the best use of the unlabeled data and enhancing the performance of the downstream tasks~\cite{gidaris2018unsupervised, zhai2019s4l}. In \cite{zhai2019s4l}, the image is rotated by a random angle, and this angle is predicted. With this auxiliary task of rotation estimation, the network makes room for performance improvement in the original task. However, the use of this auxiliary task is reported to underperform in supervised settings while it performs better in semi-supervised and self-supervised settings \cite{gidaris2018unsupervised, zhai2019s4l}.
\vspace{2mm}
\\
\noindent \textbf{Semi-Supervised Learning:} Data augmentation with consistency regularization has been popular in semi-supervised learning \cite{laine2016temporal, tarvainen2017mean, oliver2018realistic} where a set of predefined transformations are applied to the original labeled data and the outputs of the perturbed inputs are enforced to agree with the outputs of the original data~\cite{laine2016temporal}. The loss is defined as the mismatch between the outputs for the original and perturbed inputs. It is shown that consistency regularization improves robustness by smoothing the underlying data manifold \cite{oliver2018realistic}. The consistency regularization loss and the supervised loss is often aggregated. Similar ideas are also applied localization problems, and demonstrated better performance \cite{jeong2019consistency, jeong2021interpolation}. In our work, we extend this promising concept to optical flow estimation.

There have also been studies on semi-supervised optical flow estimation to reduce dependency on the labeled data. In \cite{lai2017semi}, an adversarial learning setup is used where the discriminator learns whether an optical flow is real (by comparison with the ground truth) or generated with a model. In the process of minimizing the discriminator loss, the generator with unlabeled data pairs is trained. In \cite{yan2020optical}, clean images are generated from foggy images, and foggy images are generated from clean images. A model is trained with interchangeable samples among clean and foggy images. These algorithms require additional networks to translate images into flow estimates. In our proposal, we do not require any separate network as a part of our training framework as we derive equivariance-based consistency losses simply by comparing the original pairs with the transformation pairs.

%%%%%%%%% Sec. 2

%%%%%%%%% Sec. 3
% \section{Proposed Method: S$^4$L for Optical Flow}
\section{Consistency for Optical Flow}
% \section{Proposal: Mixed Supervision (MixSup)}
\label{sec:s4l}

%\subsection{Notations}
%\label{method:notation}

Here, we summarize the notations used in this paper. We denote the ground truth optical flow as $f(I_t, I_{t+k})$ and the predicted optical flow as $\tilde{f}(I_t, I_{t+k})$ between two images $I_{t}$ and $I_{t+k}$ that are $k$ apart in time. Image size is $w \times h$. An occluded version of the original image $I_{t}$ and its corresponding occlusion mask are denoted as $I_{t,occ}$ and $O_t$, respectively. We denote the predicted occlusion mask as $\tilde{O}_t$. We also use $T(\cdot)$ and $R(\cdot)$ to denote the operations of transformation and transformation restoration, respectively.

The consistency strategies we describe below are applied in a self- and semi-supervised manner, which requires no additional ground truths. 

\subsection{Occlusion Consistency}
\label{subsec:self}

In this subsection, we discuss two techniques in our occlusion consistency strategy: zero forcing and mask match loss. 

\noindent \textbf{Zero Forcing:} In order to apply meaningful occlusions to images, we define an occlusion mask $O_t \in \mathbb{R}^{w \times h}$. We adopt the cow-mask~\cite{french2019semi, french2020milking} to create sufficiently random yet locally connected occlusion patterns as an occlusion could occur in any size, any shape, and at any position in an image while exhibiting locally explainable structures. Occlusions are mainly perpendicular to motion direction (depth discontinuities) for moving objects (camera motion) around object boundaries (scene depth discontinuities), thus occlusion regions are often connected. Using self-supervised learning with random occlusion masks enables our network to respond and learn such complex occlusion structures in the scene.

In a self-supervised manner, we apply the occlusion mask to a single image by multiplying pixel-wise the occlusion mask with the image, which allows us to obtain a new image pair ($I_t$, $I_{t,occ}$) without requiring any ground truth. Each entry of the occlusion mask $O_t$ takes a binary value; $O_t(p) = 1$ indicating a non-occluded pixel $p$ and $O_t(p)=0$ corresponds to a masked pixel. We impose the flow to be zero, i.e, $\tilde{f}(I_t, I_{t,occ})=0$, as there is no motion but only occlusion. This allow us to compute the zero-forcing loss as 
\begin{equation} 
\label{self:zero}
\begin{split}
\mathcal{L}_{ZF} = \sum_{i=1}^N \gamma^{N-i} \left\| \tilde{f^{i}}(I_t, I_{t,occ}) \right\| _1. 
\end{split}
\end{equation} 
As an enhancement to the occlusion consistency, we further introduce a special case in which $O_{t} = 1$ (no occlusion), meaning two images in the newly formed pair are identical, i.e., the pair to be $(I_t, I_t)$, which results in the new zero-forcing loss
\begin{equation} 
\label{self:zeroidentical}
% \begin{split}
\mathcal{L}_{ZF^*} = \sum_{i=1}^N \gamma^{N-i} \left\| \tilde{f^{i}}(I_t, I_t) \right\| _1.
% \end{split}
\end{equation} 

\begin{figure}[t]
\begin{center}
\includegraphics[width=0.95\linewidth]{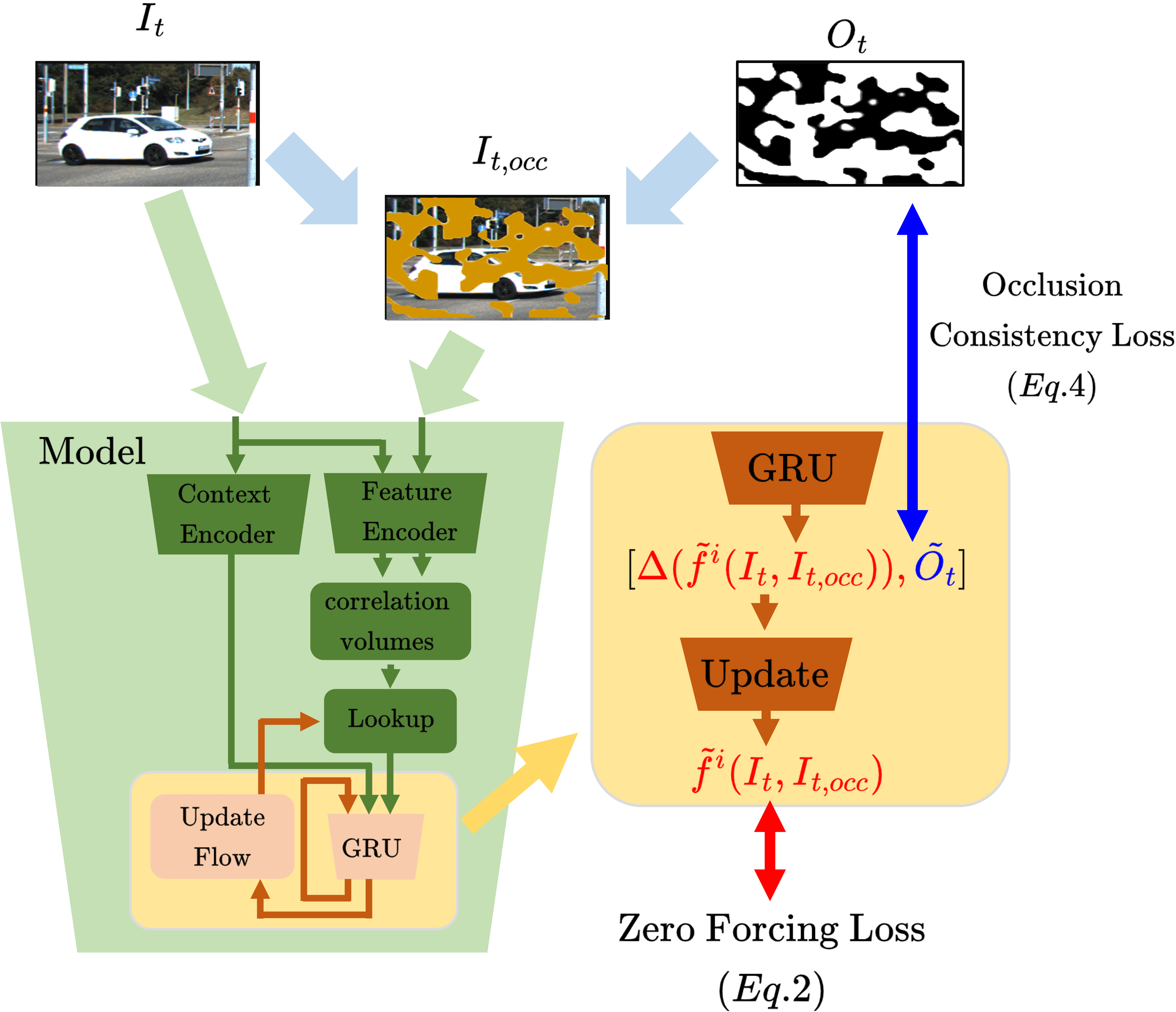} 
\vspace{-2mm}
\end{center}
\caption{\textbf{Occlusion consistency:} A random mask is applied to the original image $I_t$ to construct $I_{t,occ}$. Then, the optical flow, as well as the occlusion mask, are estimated for the image pair $(I_t, I_{t,occ})$. In this case, the target ground truth is $f(I_t,I_{t,occ}) = 0$.
% \nj{Nojun: In the figure, replacing Eq.1 with Eq.2 seems better.}
}
\vspace{-2mm}
\label{method:self}
\end{figure}

\noindent \textbf{Mask Match Loss:} Since we can generate occlusion masks automatically, our intuition is that we can also estimate them in our network and reinforce another consistency by matching the generated $O_t$ and estimated $\tilde{O}_t$ masks. To achieve this, we introduce one additional channel in the output of our network to estimate the occlusion status of pixels. This also facilitates better feature correspondences for correlation volumes as the network can directly access an internal occlusion mask in its layers. Furthermore, occlusion mask estimation can be refined over iterations and along with supervision. Therefore, we employ the zero-forcing loss together with an occlusion mask match loss simultaneously and iteratively in our occlusion consistency strategy. We define the mask match loss as
\begin{equation} 
\label{self:loss}
\begin{split}
\mathcal{L}_{MM} = \sum_{i=1}^N \gamma^{N-i} \left( - \frac{1}{wh} \sum_p O_t(p) log(\tilde{O}^i_t(p)) \right) 
\end{split}
\end{equation}
Here, we use the cross entropy, $\gamma$ and $N$ are the same parameters as defined in (\ref{related:raft}).

\subsection{Transformation Consistency}
\label{subsec:semi}

Transformation consistency strategy leverages two methods; consistency regularization and frame-hopping with semi-supervised learning.

\begin{figure}[t]
\begin{center}
\includegraphics[width=\linewidth]{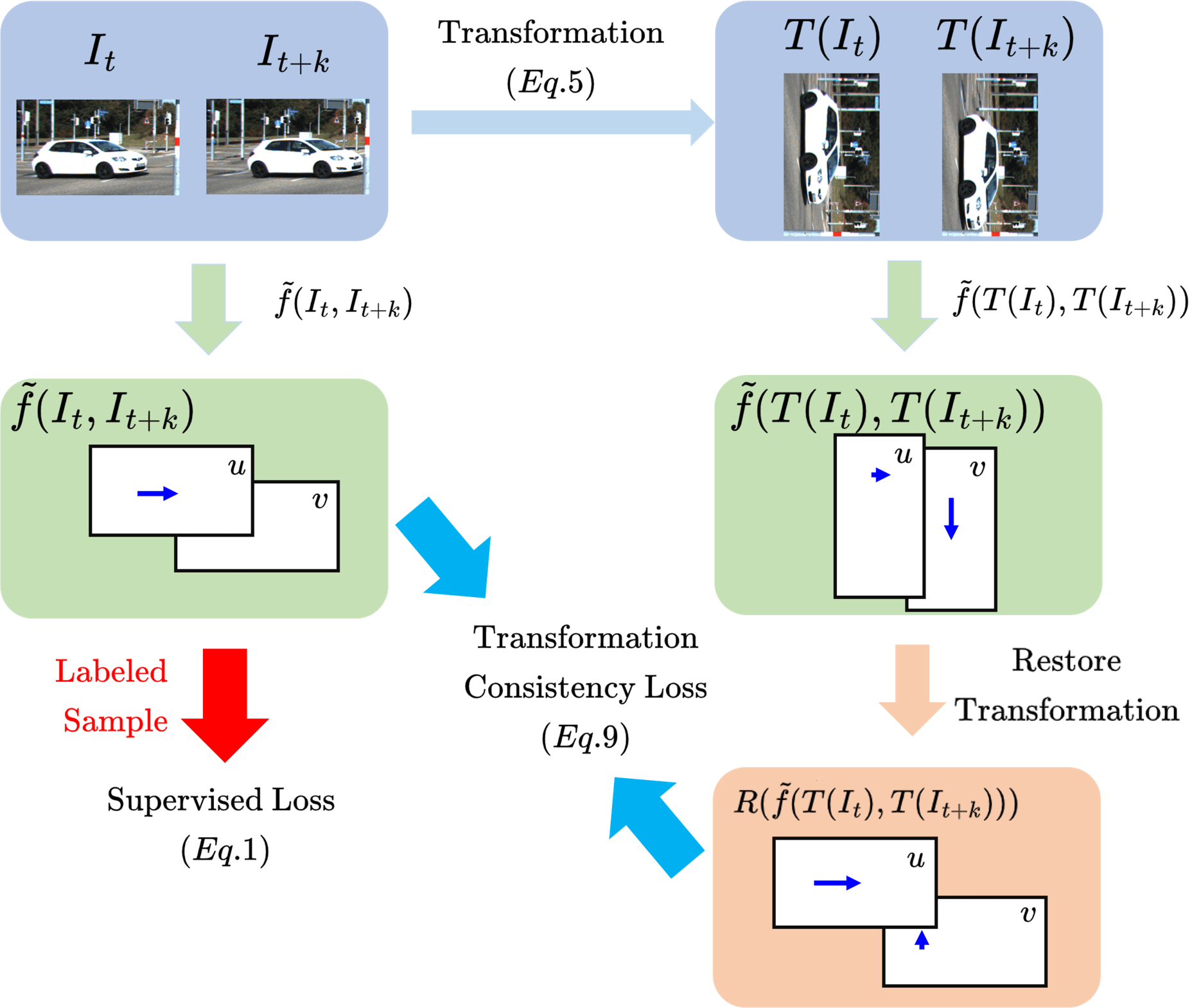}
\vspace{-7mm}
\end{center}
\caption{\textbf{Transformation consistency}. $T(I_{t})$ and $T(I_{t+k})$ are generated with image-wise transformations (random rotation as illustrated) for the image pair $(I_{t}, I_{t+k})$. Optical flows $\tilde{f}(I_{t}, I_{t+k})$ and $\tilde{f}(T(I_{t}), T(I_{t+k}))$ are computed by the same model for the image pair and its transformed image pair. Then, the estimated flow for the transformed pair are remapped by applying the transformation restoration operation.
%\js{restore} transformation. 
In case we have labeled data, a supervised loss is calculated between $\tilde{f}$ and the ground truth $f$. 
}
% \vspace{-2mm}
\label{method:semi}
\end{figure}

We apply spatial transformation consistency to the input image pair, creating cases for enforcing equivariance between the estimated optical flow for the original pair and the estimated optical flow for the transformed pair, in addition to the supervised loss of optical flow (See Fig.~\ref{method:semi}). In addition, as an enhancement to this transformation consistency methods, we extend the temporal gap from $k=1$ to $k \ge 1$ to include pairs where the images depict larger motions. Existing datasets typically provide ground truth flow fields $f(I_t,I_{t+1})$ only between consecutive image frames $I_{t}$ and $I_{t+1}$, while the image sampling rates may vary\footnote{For example, the frame rate of the Sintel~\cite{butler2012naturalistic} dataset is 24 frames-per-second, while that of the KITTI~\cite{geiger2013vision} is 10 frames-per-second.} significantly from one dataset to another. Allowing pairs with larger frame gaps enables more versatile characterization of underlying object and camera motion with different speeds.

% \subsubsection{Consistency Regularization with Image-wise Transformation} 
\noindent \textbf{Consistency Regularization:} Optical flow estimations should equivariantly change when the input images in the pair undergo the same spatial (geometric) transformations that are bijective. We take advantage of this property and impose an intuitive consistency regularization for the image pairs during the training process. More specifically, we apply 2D image transformations, including flips and random rotations that we observed to be effective choices, to the input images and corresponding estimated optical flows.  

Figure~\ref{method:semi} shows an example for the transformation consistency regularization. We transform both images $I_{t}$ and $I_{t+k}$ in the pair
\begin{equation} 
\label{semi:image}
% \begin{split}
I_{t}, I_{t+k} \xmapsto{T} T(I_{t}), T(I_{t+k})
% I_{t}, I_{t+k} \Longrightarrow R(I_{t}), R(I_{t+k})
% \end{split}
\end{equation}
and compute the optical flow for the original and transformed pairs using our model. Our assumption is that after applying transformation restoration, the estimated optical flows should be equivalent
\begin{equation} 
\label{semi:each_output}
\tilde{f}(I_{t}, I_{t+k}) = R\left( \tilde{f}(T(I_{t}), T(I_{t+k})) \right).
% \tilde{f}(I_{t}, I_{t+k}) = T^{-1}\left( \tilde{f}(T(I_{t}), T(I_{t+k})) \right).
\end{equation}
Using this, we compute the transformation consistency loss $\mathcal{L}_{tr}$ between $\tilde{f}$ and $R(\tilde{f})$ as follow
\begin{equation} 
% \begin{split}
\label{eq:semi-loss}
\mathcal{L}_{tr} = \left\| \tilde{f}(I_{t}, I_{t+k})  - R\left( \tilde{f}(T(I_{t}), T(I_{t+k})) \right) \right\|_{2}^{2}.\\
% l_{u} = \lVert \tilde{V} - R^{-1}(\tilde{R(V))} \lVert _{2}^{2}.
% \end{split}
\end{equation}

During the initial phase of training, a larger transformation inconsistency $\mathcal{L}_{tr}$ is more likely to occur, thus the training may diverge. To alleviate this issue, we introduce an \textit{identifier mask} $\alpha$ ($\alpha \in$ $\mathbb{R}^{w \times h}$) as follows
\begin{equation} \label{eq:eb}
\alpha^{i} = \begin{cases}
1, & \mbox{if} \ \mathcal{L}^{i}_{tr} < \epsilon \\
0, & \mbox{otherwise}.
\end{cases}
\end{equation}
Here, $\epsilon$ is a small positive constant, which is then used in the final loss function to prevent the network from diverging
\begin{equation} \label{eq:L-semi}
\mathcal{L}_{TR} =\sum_{i=1}^N \gamma^{N-i} \cdot  \mathbb{E}_{\mathbb{I}\{\alpha^{i} = 1\}}(\mathcal{L}^i_{tr}).
\end{equation}
where $\mathbb{I}\{\alpha^{i} = 1\}$ indicates that the expectation is fulfilled only for the ones in mask.
For iterative flow refinement, $\mathcal{L}^i_{tr}$ is calculated in the $i$-th iteration as in (\ref{eq:semi-loss}) and $\gamma$ and $N$ are the same parameters as (\ref{related:raft}). 

\noindent \textbf{Frame Hopping: } We also utilize \textit{frame hopping}, a technique inspired by ScopeFlow \cite{bar2020scopeflow}. Our intuition is that larger displacements in the datasets \cite{butler2012naturalistic, geiger2013vision} exist mostly near edges of images; thus, training with samples containing larger displacements can benefit model performance. Frame hopping (for image pairs $(I_{t}, I_{t+k})$ with $k>1$) provides not only more training samples but also samples with larger displacements to enhance learning.

\begin{table*}[t]
\begin{center}
\caption{Optical Flow results for Sintel and KITTI. We trained the model with the Flyingchairs (C) and Flyingthings (T) datasets and tested the model on the training dataset of the Sintel (S) and KITTI (T).
For Sintel and KITTI tests, we finetuned the model with a pre-trained model (C+T) with the Sintel, KITTI, and HD1K (H) training dataset. (Smaller numbers are better. The numbers in gray have little meaning because they are measured on the training data. $\dagger$ is trained including test images without label as unlabeled data, and $\ddagger$ is trained on KITTI-2012 and KITTI-2015 datasets. 
* is the results of warm-start, and $\S$ is the results of undisclosed method.
) }
\label{tab:expsem2}
\vspace{-2mm}
%\footnotesize
\adjustbox{max width=0.98\textwidth}
{
\begin{tabular}{|l||c||c|c||c|c||c|c||c|}
\hline
% Method & Labeled & Network & \multicolumn{2}{|c|}{mAP (\%)}\\
\multirow{2}{*}{Method} & Training  & \multicolumn{2}{|c|}{Sintel (train-EPE)}  & \multicolumn{2}{|c|}{KITTI (train)} & \multicolumn{2}{|c|}{Sintel (test-EPE)}  & KITTI (test) \\
\cline{3-9}
&  dataset & (Clean) & (Final) & (Fl-epe) & (Fl-all) & (Clean) & (Final) & (Fl-all) \\
\hline
\hline
HD3 \cite{yin2019hierarchical} & \multirow{11}{*}{C+T} & 3.84 & 8.77 & 13.17 & 24.0 & - & - & - \\
FlowNet2 \cite{ilg2017flownet} & & 2.02 & 3.54 & 10.08& 30.0 & 3.96 & 6.02& - \\
PWC-Net \cite{sun2018pwc} &  & 2.55 & 3.93 & 10.35 & 33.7 & - & - & - \\
LightFlowNet \cite{hui2018liteflownet} &  & 2.48 & 4.04 & 10.39 & 28.5 & - & - & - \\
LightFlowNet2 \cite{hui2019lightweight} &  & 2.24 & 3.78 & 8.97 & 25.9 & - & - & - \\
VCN \cite{yang2019volumetric} & & 2.21 & 3.68 & 8.36 & 25.1 & - & - & - \\
MaskFlowNet \cite{zhao2020maskflownet} &  & 2.25 & 3.61 & - & 23.1 & - & - & - \\
% \hline
% \hline
\cline{1-1}
\cline{3-9}
RAFT-small \cite{teed2020raft} &  & 2.21 & 3.35 & 7.51 & 26.9 & - & - & -  \\
\textbf{Ours (RAFT-small + OCTC)} &  & \textbf{1.95} &\textbf{3.13} & \textbf{6.53} & \textbf{22.1} & - & - & - \\
\cline{1-1}
\cline{3-9}
RAFT \cite{teed2020raft}      & & 1.43 & 2.71 & 5.04 & 17.4 & - & -  & - \\
\textbf{Ours (RAFT + OCTC)}&  & \textbf{1.31} & \textbf{2.67} & \textbf{4.72} & \textbf{16.3} & - & -  & - \\

\hline
\hline
SelFlow \cite{liu2019selflow} & \multirow{2}{*}{C+T+S+K} & \textcolor{gray}{1.68} & \textcolor{gray}{1.77} & \textcolor{gray}{-} & \textcolor{gray}{1.18} & 3.74 & 4.26 & 8.42 \\
ScopeFlow \cite{bar2020scopeflow} &  & \textcolor{gray}{-} & \textcolor{gray}{-} & \textcolor{gray}{-} & \textcolor{gray}{-} & 3.59 & 4.10 & 6.82 \\

\hline
LiteFlowNet2 \cite{yin2019hierarchical} & \multirow{5}{*}{C+T+S+K+H} & \textcolor{gray}{1.30} & \textcolor{gray}{1.62} & \textcolor{gray}{1.47} & \textcolor{gray}{4.8} & 3.48 & 4.69 & 	7.62 \\
PWC-Net+ \cite{sun2019models} & & 
\textcolor{gray}{1.71} & \textcolor{gray}{2.34} & \textcolor{gray}{1.50} & \textcolor{gray}{5.3} & 3.45 & 4.60 & 7.72 \\
VCN \cite{yang2019volumetric} & & 
\textcolor{gray}{1.66} & \textcolor{gray}{2.24} & \textcolor{gray}{1.16} & \textcolor{gray}{4.1} & 2.81 & 4.40 & 6.30 \\
MaskFlowNet \cite{zhao2020maskflownet} & &
\textcolor{gray}{-} & \textcolor{gray}{-} & \textcolor{gray}{-} & \textcolor{gray}{-} & 2.52 & 4.17 & 6.10 \\
%\cline{1-1}
%\cline{3-10}
RAFT \cite{teed2020raft} & &
\textcolor{gray}{0.76} & \textcolor{gray}{1.22} & \textcolor{gray}{0.63} & \textcolor{gray}{1.5} & 1.94/1.61* & 3.18/2.86* & 5.10 \\

\cline{2-2}

CRAFT \cite{craft2021} & Undisclosed &
\textcolor{gray}{-}  & \textcolor{gray}{-} & \textcolor{gray}{-} & \textcolor{gray}{-} & $1.45^{\S}$ & $2.42^{\S}$ & 4.79 \\

RAFT-A \cite{sun2021autoflow} & A+T+S+K+H &
-  & - & - & - & 2.01/\enspace --\enspace* & 3.14/\enspace --\enspace* & 4.78 \\

\cline{2-2}

GMA \cite{jiang2021learning} & \multirow{4}{*}{C+T+S+K+H} &
\textcolor{gray}{0.62}  & \textcolor{gray}{1.06} & \textcolor{gray}{0.57} & \textcolor{gray}{1.2} & \enspace --\enspace \enspace /\textbf{1.39}* & \enspace --\enspace \enspace /\textbf{2.47}* & 5.15 \\

%\cline{1-1} \cline{3-9}

\textbf{Ours (RAFT + OCTC)} & & 
\textcolor{gray}{0.73} & \textcolor{gray}{1.23} & \textcolor{gray}{0.67} & \textcolor{gray}{1.7} &
1.82/\enspace --\enspace* & 3.09/\enspace --\enspace*  & 4.72  \\

\textbf{Ours$^{\dagger}$ (RAFT + OCTC)} & & 
\textcolor{gray}{0.74} & \textcolor{gray}{1.24} & \textcolor{gray}{0.71} & \textcolor{gray}{2.0} & 1.58/\enspace --\enspace* & \textbf{2.95}/\enspace --\enspace*  & \enspace --\enspace \\ 

\textbf{Ours$^{\ddagger}$ (RAFT + OCTC)} & & 
\textcolor{gray}{-} & \textcolor{gray}{-} & \textcolor{gray}{0.78} & \textcolor{gray}{ 2.3} &
\textbf{1.55}/1.41* &  2.98/2.57* & \textbf{4.33} \\ 

\hline
\end{tabular}
}
\vspace{-5mm}
\end{center}
\end{table*}

% \vspace{-5mm}
\subsection{Aggregated Loss}
% \subsection{Mixed Supervision for Optical Flow}
\label{subsec:semiself}

Our total loss consists of the conventional supervised loss ($\mathcal{L}_{base}$), the zero-forcing loss ($\mathcal{L}_{ZF}$), the mask match loss ($\mathcal{L}_{MM}$), and the transformation consistency loss ($\mathcal{L}_{TR}$) as follows:
\begin{equation} 
\label{total:loss}
\mathcal{L}_{total} = \mathcal{L}_{base} + \mathcal{L}_{ZF} + \lambda_1 \mathcal{L}_{MM} + \lambda_{2} \mathcal{L}_{TR}.
\end{equation}
The supervised loss ($L_{base}$ in (\ref{related:raft})) for labeled data and the unsupervised loss ($L_{ZF}$ in (\ref{self:zero})), ($L_{MM}$ in (\ref{self:loss})), and ($L_{TR}$ in (\ref{eq:L-semi})) for unlabeled data are combined by using a balance parameter $\lambda_1$ and $\lambda_2$ to derive the final loss\footnote{Zero Forcing loss is computed with the same balance with supervised learning.}.

\section{Experiments}
\label{sec:exp}

\noindent \textbf{Datasets \& Implementation Details:}
In our experiments, we have utilized the FlyingChairs (C) \cite{dosovitskiy2015flownet},
FlyingThings3D (T) \cite{mayer2016large},
Sintel (S) \cite{butler2012naturalistic}, KITTI (K) \cite{geiger2013vision, menze2015object}, and HD1K(H) \cite{kondermann2016hci} datasets, which are the most popular benchmarks in the optical flow estimation problem. More details on our experimental analysis are provided in the supplementary material.

All experiments have been conducted under the same setting with the official code of RAFT\footnote{https://github.com/princeton-vl/RAFT}. We followed the same batch sizes, optimizer, number of GRU iterations, and so on. As the number of image pairs increased in our method, we increased the number of iterations proportionally. Similar to RAFT, we pretrained our model in sequence with FlyingChairs and FlyingThings3D. Since Flyingchair samples do not have more than two consecutive images, only self-supervised learning was applied. The parameters are set to ($\lambda_{1}$, $\lambda_{2}$) = (0.1, 0.01) in (\ref{total:loss}), $\epsilon$ = $5^{2}$ in (\ref{eq:eb}), and $k$ is set to 2. \footnote{We performed a grid search in \{$3^{2}$, $5^{2}$, $7^{2}$, $\infty$\} for $\epsilon$ value in Eq.\ref{eq:eb} and over the values in \{1.0, 0.1, 0.01, 0.001\} for each $\lambda$ in Eq.\ref{total:loss}. The best hyperparameters found were [$\epsilon$ = $5^{2}$, ($\lambda_{1}$, $\lambda_{2}$) = (0.1, 0.01)]. More details and results of these experiments are provided in Supplementary File.} For a wide variety of random patterns in occlusion consistency learning, we applied cowmask\footnote{\url{https://github.com/google-research/google-research/tree/master/milking_cowmask}} with the same parameters used in \cite{french2020milking}. All samples applied in our experiments are from the original datasets without additional data.
\\
\begin{figure*}[t]
\begin{center}$
\centering
\begin{tabular}{c c c c c}

\textbf{\small{Ground Truth}} & \textbf{RAFT} & \textbf{Ours} & \textbf{\small{Occlusion GT}} & \textbf{Ours} ($\tilde{O_{t}}$) 
% \vspace{0.2cm}
\\

\hspace{-0.2cm} \includegraphics[width=3.4cm, height=1.6cm]{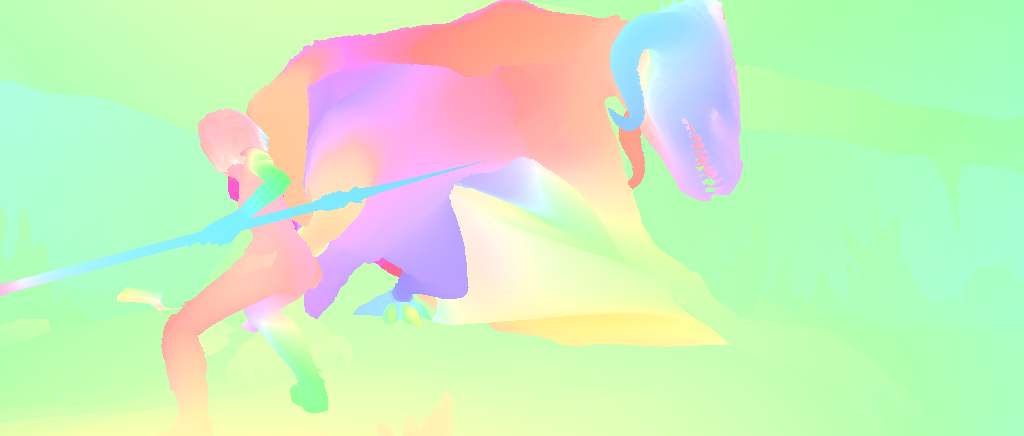} & \hspace{-0.4cm}
\includegraphics[width=3.4cm, height=1.6cm]{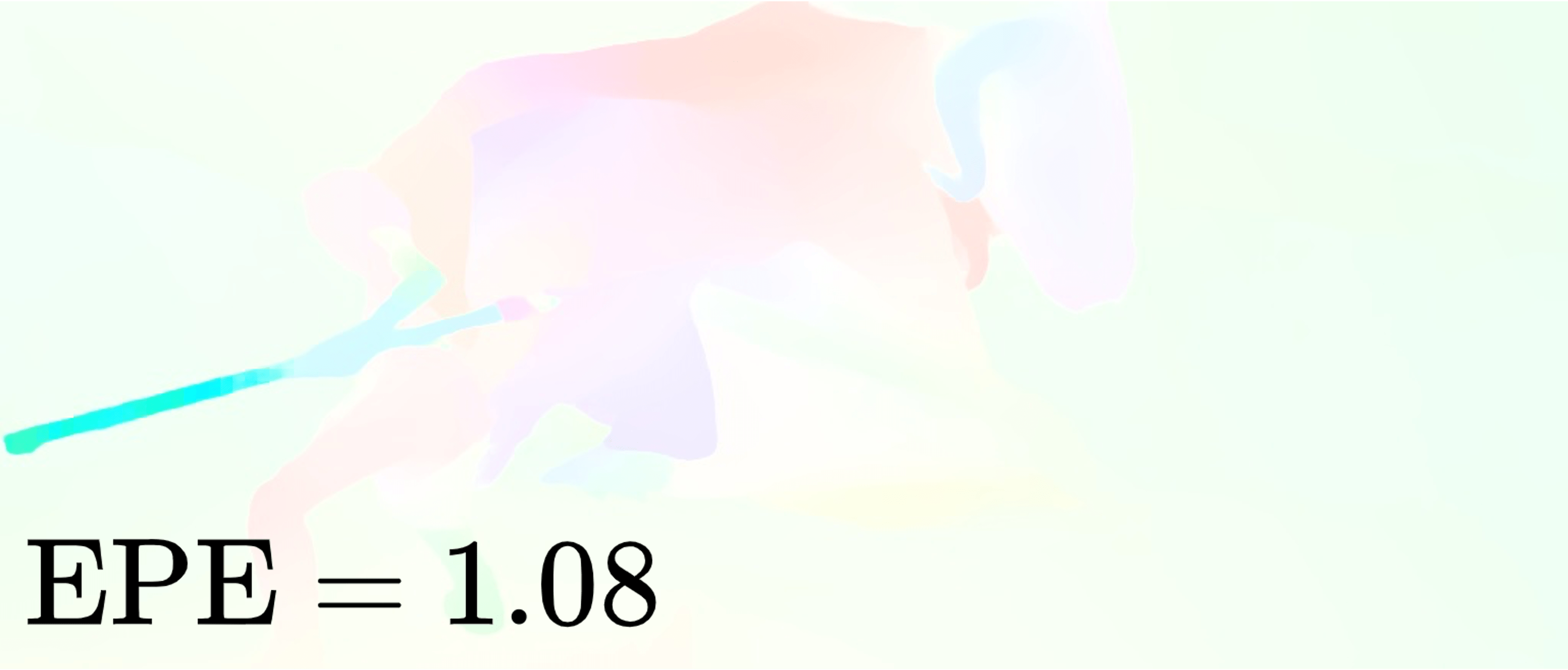} & \hspace{-0.4cm}
\includegraphics[width=3.4cm, height=1.6cm]{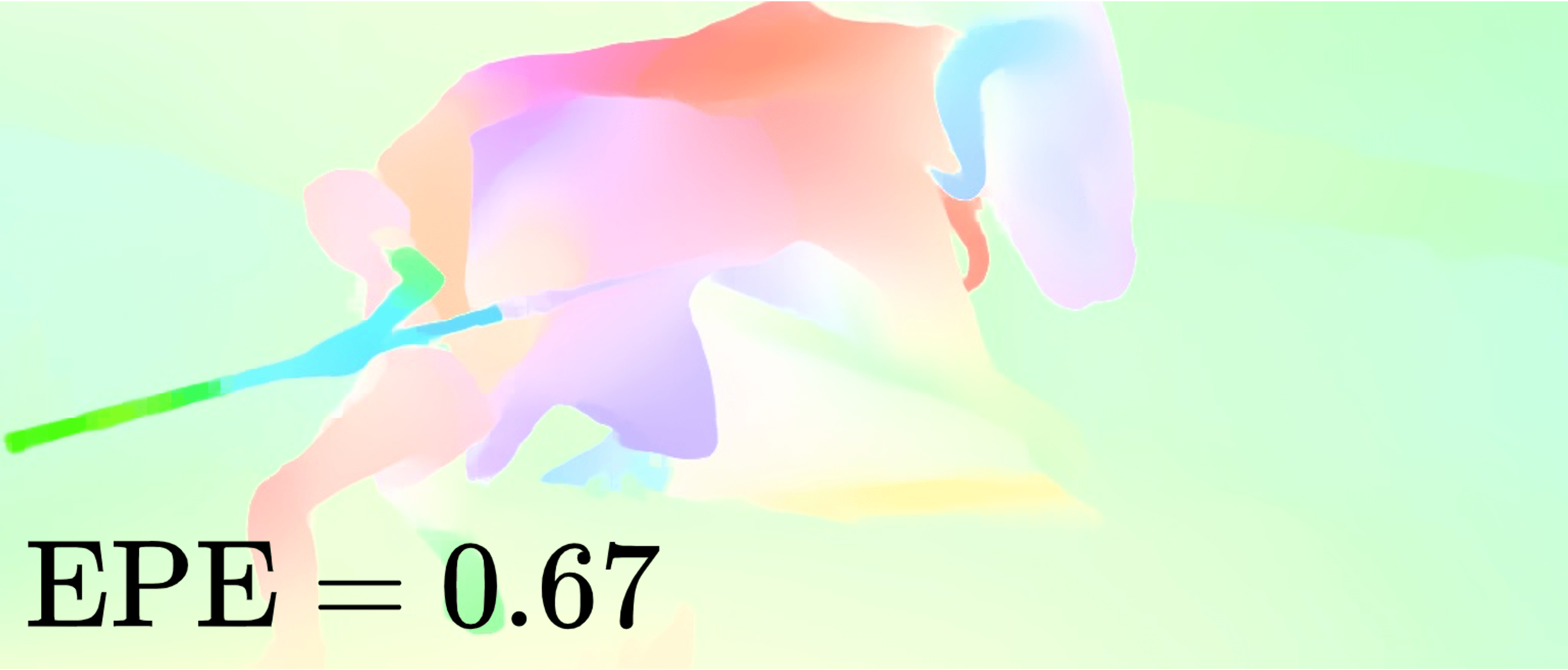} & \hspace{-0.4cm}
\includegraphics[width=3.4cm, height=1.6cm]{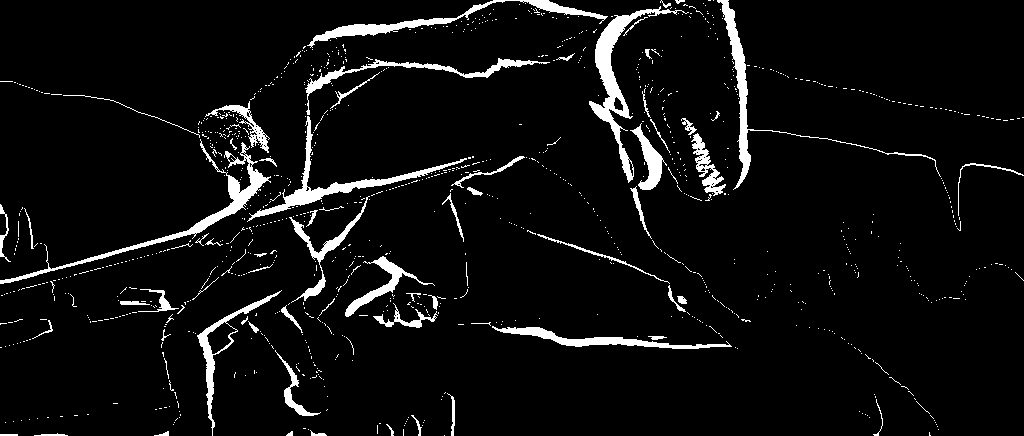} & \hspace{-0.4cm}
\includegraphics[width=3.4cm, height=1.6cm]{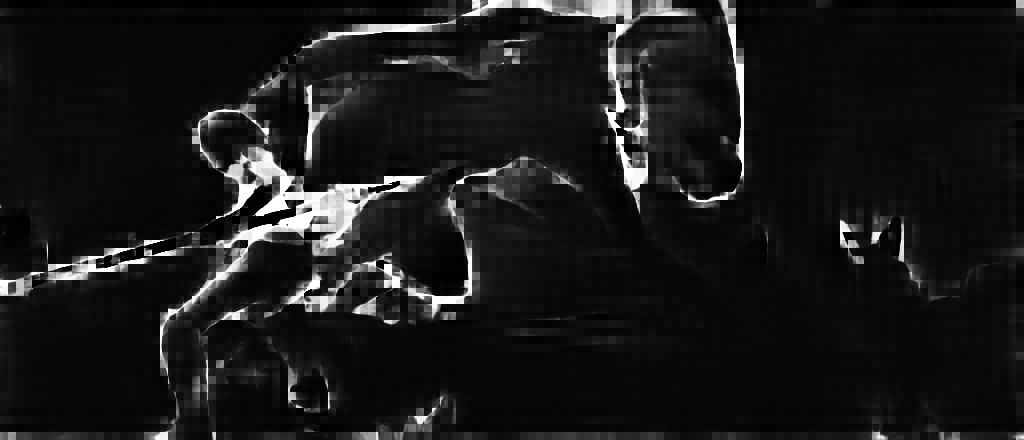}  \\

\hspace{-0.2cm} \includegraphics[width=3.4cm, height=1.6cm]{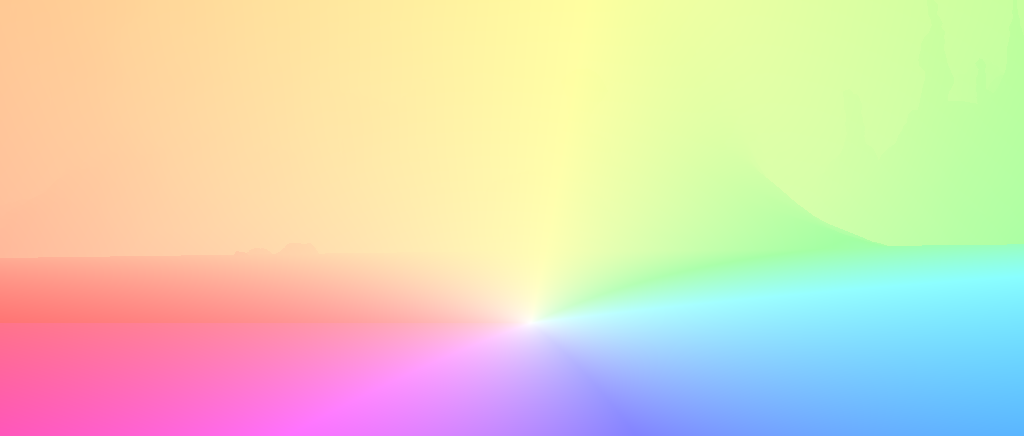} & \hspace{-0.4cm}
\includegraphics[width=3.4cm, height=1.6cm]{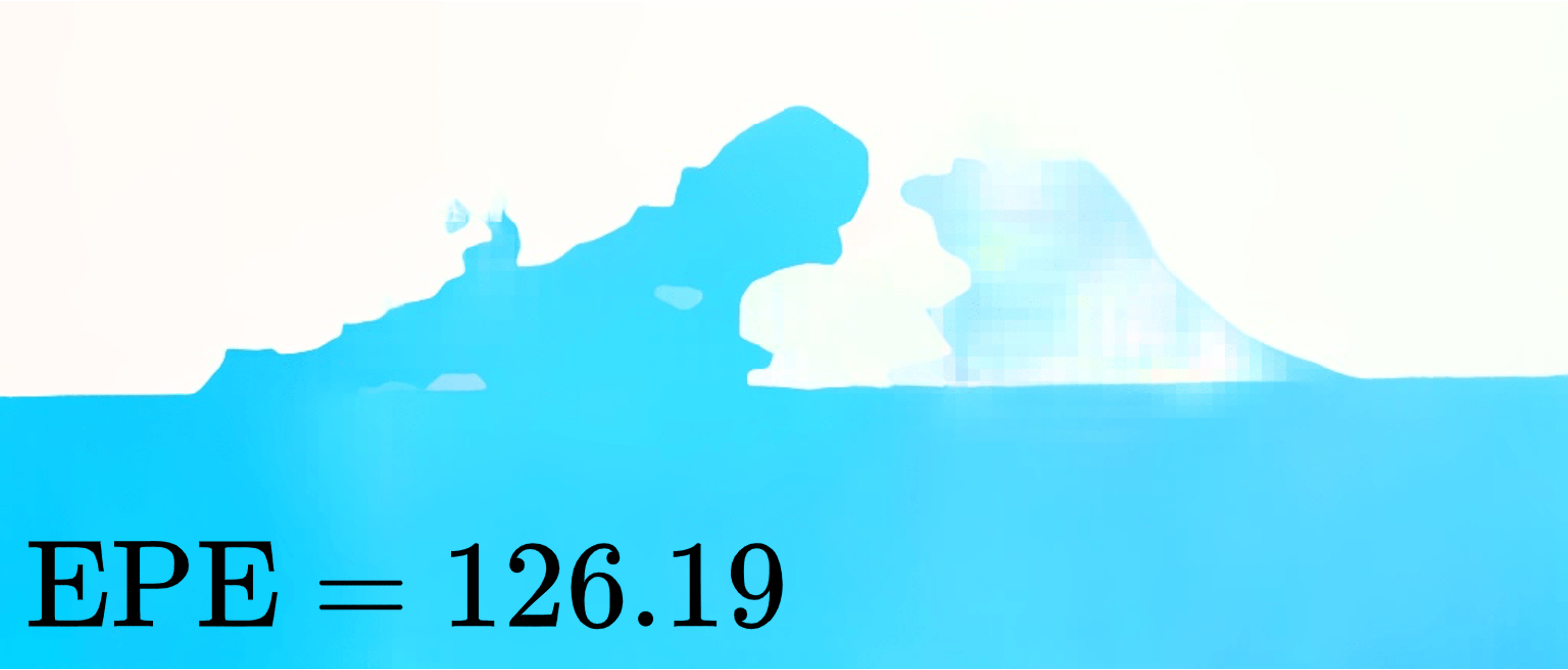} & \hspace{-0.4cm}
\includegraphics[width=3.4cm, height=1.6cm]{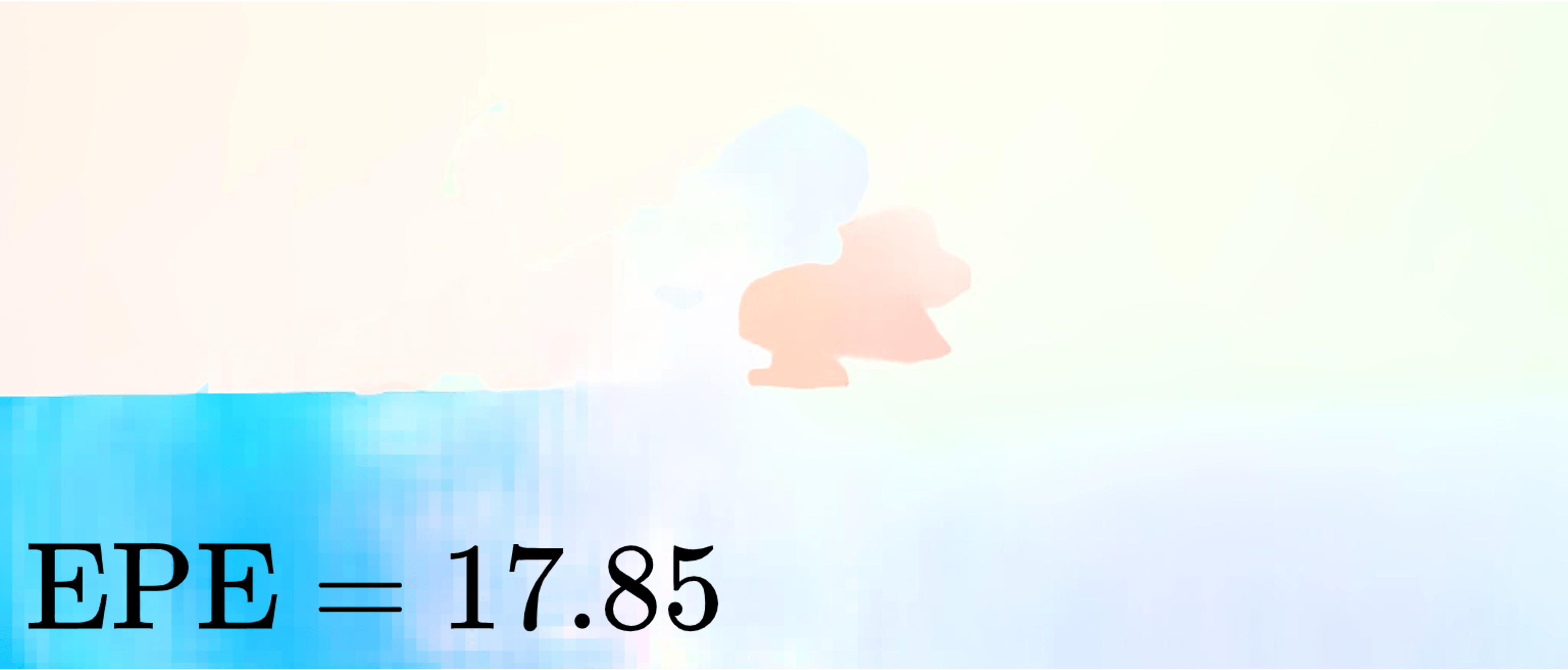} & \hspace{-0.4cm}
\includegraphics[width=3.4cm, height=1.6cm]{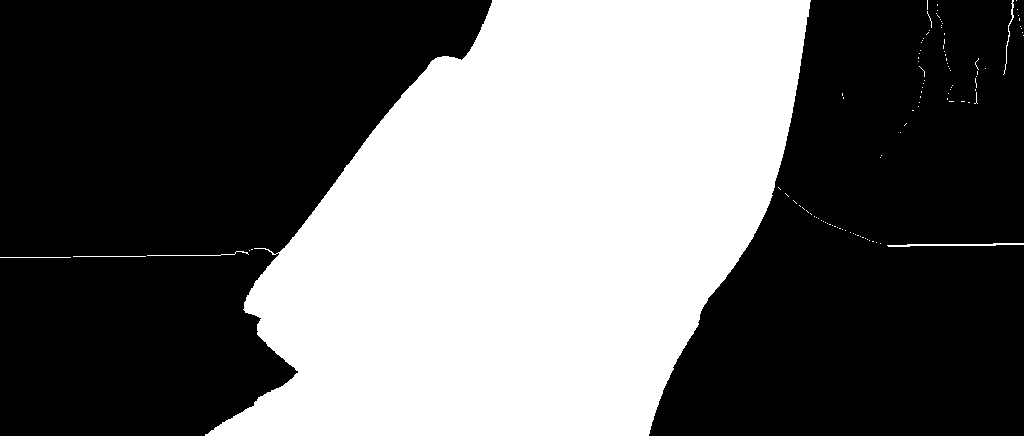} & \hspace{-0.4cm}
\includegraphics[width=3.4cm, height=1.6cm]{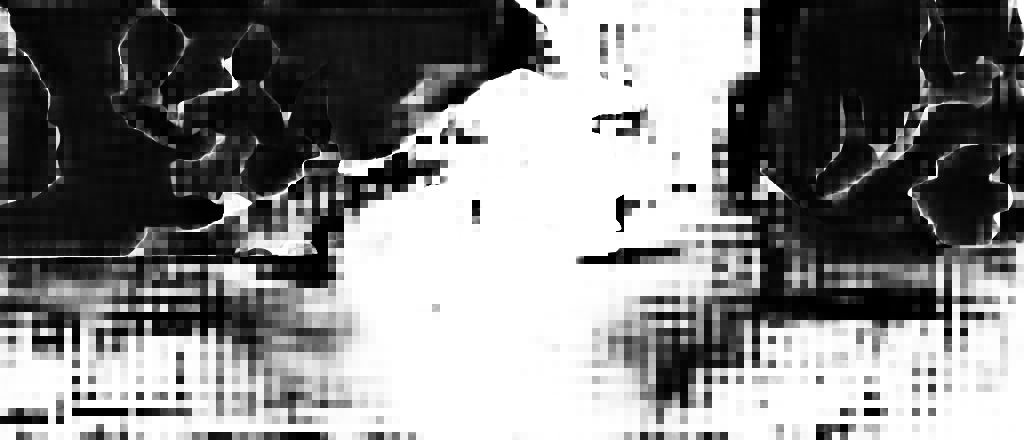}  

\vspace{-4mm}
\end{tabular}$
\end{center}
% \vspace{-2mm}
\caption{Qualitative results for the Sintel training set using RAFT and our RAFT+OCTC (Occlusion Consistency and Transformation Consistency) models (trained with C+T). The first row shows that our RAFT+OCTC, which adopts frame hopping in transformation consistency, works better for large displacements than RAFT. The second row shows that our RAFT+OCTC can predict occlusion area, and it helps our model prevent incorrect predictions. 
% \vspace{-3mm}
}
\label{exp:example_image}
% \vspace{-1mm}
\end{figure*}
\begin{figure*}[t]
\begin{center}$
\centering
\begin{tabular}{c c c}

\textbf{Image} & \textbf{RAFT} & \textbf{Ours}
% \vspace{0.2cm}
\\ 

\hspace{-0.2cm} \includegraphics[width=5.7cm, height=1.7cm]{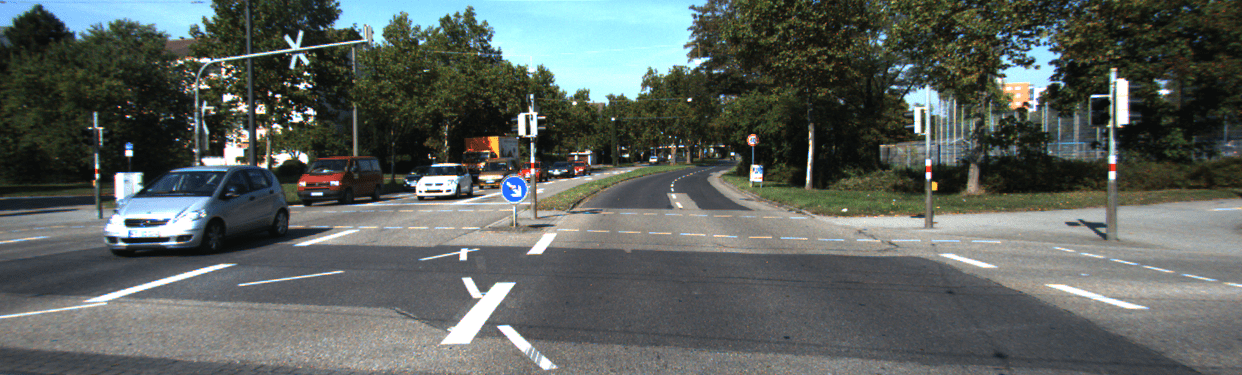} & \hspace{-0.4cm}
\includegraphics[width=5.7cm, height=1.7cm]{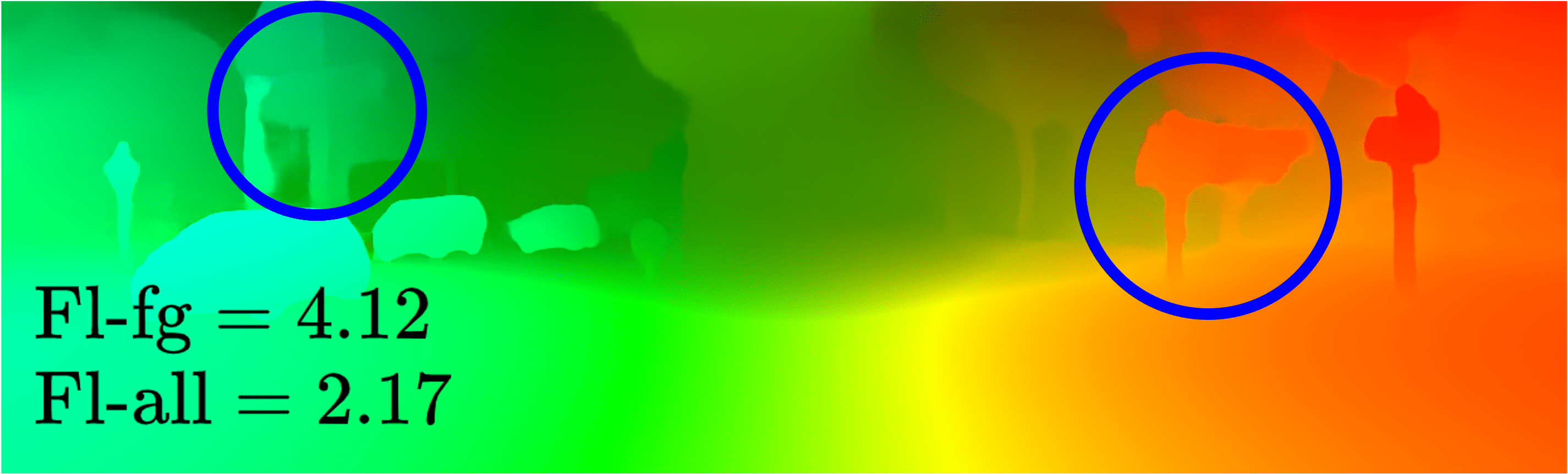} & \hspace{-0.4cm}
\includegraphics[width=5.7cm, height=1.7cm]{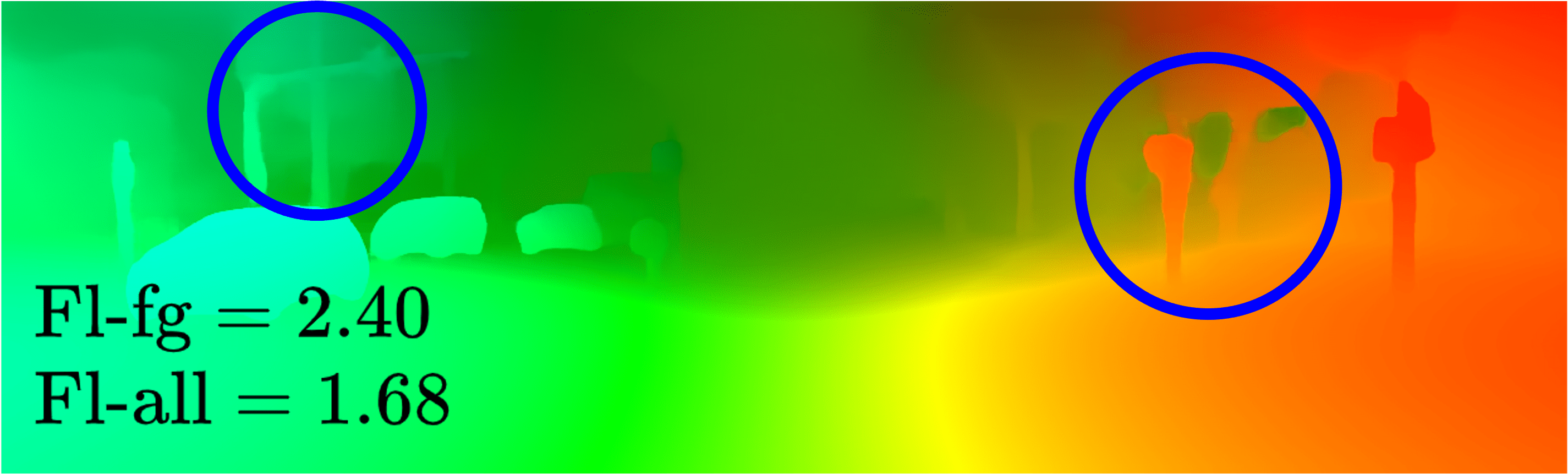} \\

\hspace{-0.2cm} \includegraphics[width=5.7cm, height=1.7cm]{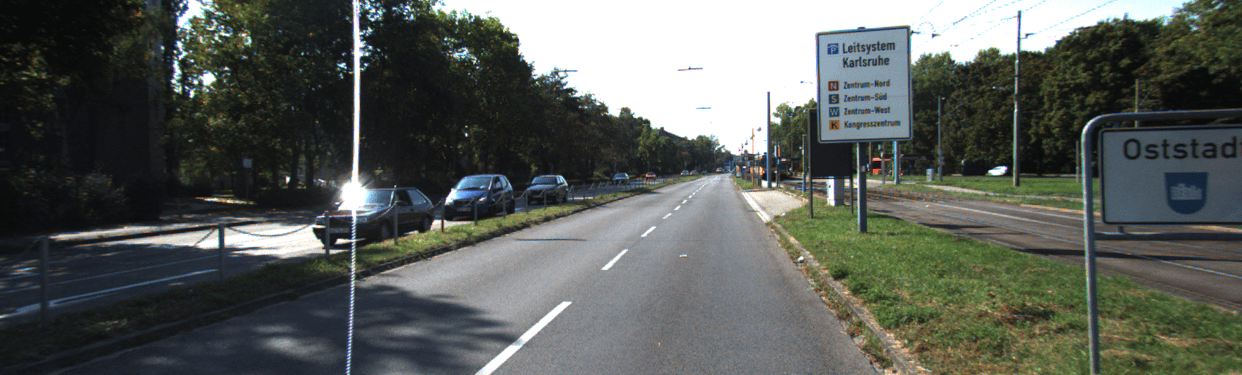} & \hspace{-0.4cm}
\includegraphics[width=5.7cm, height=1.7cm]{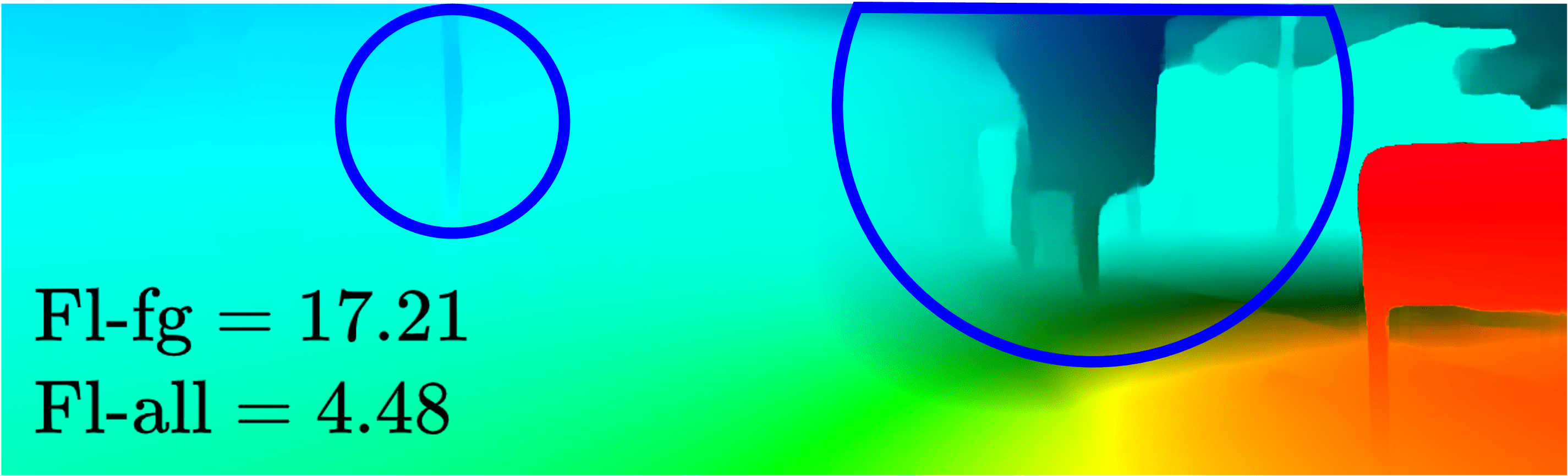} & \hspace{-0.4cm}
\includegraphics[width=5.7cm, height=1.7cm]{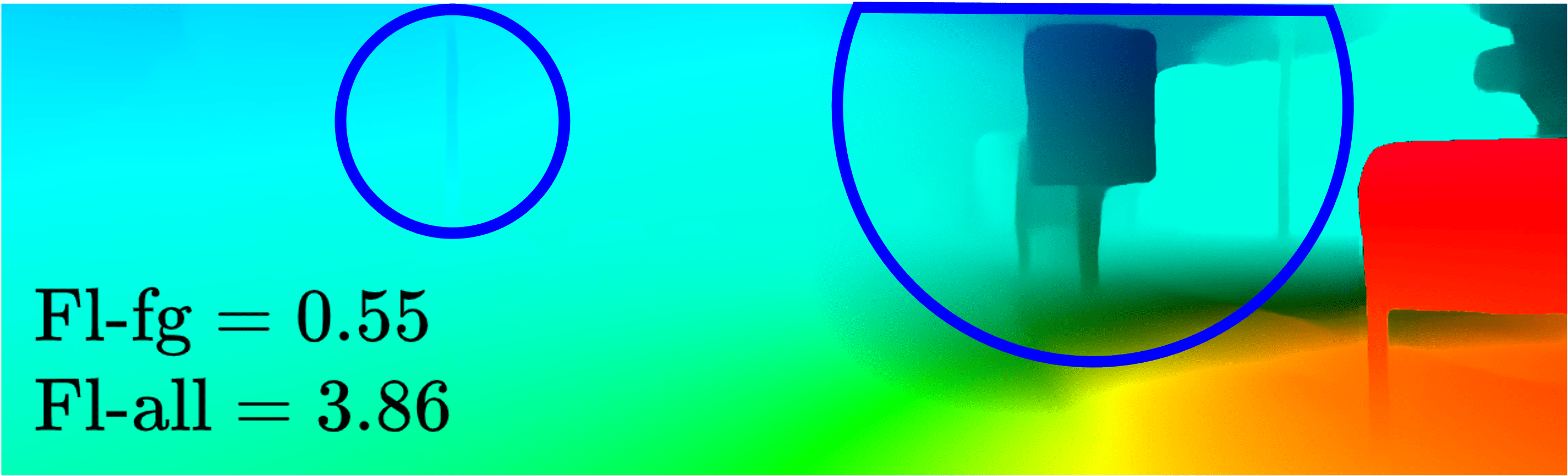}

\vspace{-4mm}
\end{tabular}$
\end{center}
% \vspace{-2mm}
\caption{Qualitative results for the KITTI test set using RAFT and our RAFT+OCTC (Occlusion Consistency and Transformation Consistency) models (trained with C+T+S+K+H). 
}
\vspace{-3mm}
\label{exp:example_image_kitti}
\end{figure*}

\noindent 
\textbf{Experimental Results: }
Table~\ref{tab:expsem2} shows the performances of the proposed method and some very recent optical flow estimation algorithms. The model trained with C+T, RAFT reported the state-of-the-art performance previously. Nevertheless, we improved its performance even further when we applied our learning scheme OCTC (Occlusion Consistency and Transformation Consistency). In addition, our method outperformed others on the KITTI benchmark that contains real images. Our method achieved 0.26 and 0.22 EPE improvements in Sintel-clean and Sintel-final, respectively, in relation to RAFT-small. For the KITTI dataset, EPE decreased by an impressive 0.98, and Fl-all decreased by 4.8\%. Using the RAFT-large model, our performance in predicting the optical flow still attained additional improvements; 0.12 and 0.04 smaller EPE for Sintel-clean and Sintel-final, and 0.32 EPE decrease and 1.1\% Fl-all decrease for the KITTI dataset.

The bottom half of Table~\ref{tab:expsem2} presents the performance on the test datasets of Sintel and KITTI. The models are trained with the training datasets of Sintel and KITTI. For the model trained on the Sintel dataset, the test EPE decreased by 0.12 and 0.09 for clean and final, respectively, compared to RAFT. For the model trained on the KITTI-2015 dataset, the Fl-all score our model improves down to 4.72\%. Furthermore, we trained our model with test images without labels treating them as unlabeled data. In Sintel, the test EPEs are 1.58 and 2.95 in the clean and final versions, respectively. Like MaskFlowNet, when we finetune on KITTI-2012 and KITTI-2015 together, our model shows further performance improvement with an Fl-all score of 4.33\%, which achieves the new state of the art on the KITTI-2015 dataset. The proposed method has a gain of about 0.77\% over the conventional RAFT model. And, when we applied our method with warm-start, it also shows the performance improvement.

In comparison to other algorithms, our method brings robust improvements for both the Sintel and KITTI datasets. RAFT-A~\cite{sun2021autoflow} shows performance improvement in the KITTI dataset, but its performance degrades in the Sintel dataset. GMA~\cite{jiang2021learning} reports state-of-the-art performance in the Sintel dataset, but its performance is not consistent; it is worse than the baseline RAFT in the KITTI dataset.

\noindent \textbf{Qualitative Results: }
Figure~\ref{exp:example_image} provides qualitative comparisons on the Sintel training dataset, where the scenarios of long-range movement and of large-area occlusion are shown in the top and bottom rows, respectively. In both scenarios, our model demonstrates improved accuracy than the RAFT baseline, indicating the effects of our consistency imposing strategies. Specifically, in the top row, our model trained with frame hopping enables improved handling with longer-range motions. In the bottom row, our RAFT+OCTC demonstrates improved robustness with large-area occlusions (see Supplementary file for more examples).

Figure~\ref{exp:example_image_kitti} provides qualitative comparisons on the KITTI test dataset, where our algorithm also demonstrates improved consistency in the prediction outputs.

% ($L_{ZF}$ in (\ref{self:zero})), ($L_{MM}$ in (\ref{self:loss}))
\begin{table}[t]
% \begin{subtable}
\centering
\caption{Ablation study for Occlusion Consistency (OC). We trained our models with the Flyingchairs (C) and Flyingthings (T) datasets and tested on the training dataset of the Sintel (S) and KITTI (T). $L_{ZF}$ and $L_{MM}$ are zero-forcing loss in (\ref{self:zero}) and mask match loss in (\ref{self:loss}), respectively.}
\vspace{-2mm}
\label{tab:abl2}
% \scriptsize
\adjustbox{max width=0.47\textwidth}
{
\begin{tabular}{|@{}l@{}|@{}c@{}||c|@{}c@{}||c|@{}c@{}|}
\hline
\multirow{2}{*}{Method (small)} & 
Additional & \multicolumn{2}{|@{}c@{}||}{Sintel (train-EPE)}  & \multicolumn{2}{|@{}c@{}|}{KITTI-15 (train)} \\
\cline{3-6}
& Loss &  Clean & Final & Fl-epe & Fl-all \\
\hline
RAFT (baseline) & - & 2.21 & 3.35 & 7.51 & 26.9\\
\hline
\multirow{4}{*}{RAFT + OC} & $\mathcal{L}_{ZF^*}$ ($I_{t}$,$I_{t}$) & 2.23 & 3.59 & 8.27 & 25.8\\
 & $\mathcal{L}_{ZF}$ ($I_{t}$,$I_{t,occ}$) & 2.17 & 3.35 & 7.22 & 24.2\\
 & $L_{MM}$ & 2.11 & 3.31 & 7.14 & 24.3 \\
 & $\mathcal{L}_{ZF}$ ($I_{t}$,$I_{t,occ}$) + $L_{MM}$ & \textbf{2.05} & \textbf{3.18} & \textbf{7.07} & \textbf{23.5}\\
\hline
\end{tabular}
}
\centering
\vspace{-2mm}
\end{table}

% \section{Ablation Studies}
\section{Discussion}
\label{sec:dis}
\noindent \textbf{Occlusion Consistency Terms: }
\label{dis:self_matching}
As shown in Table~\ref{tab:abl2}, when we initially used ($I_t$, $I_t$) for zero forcing (i.e., identical samples as a special case without occlusions), we observed a performance degradation possibly due to overfitting. As we applied occlusions in one of the samples ($I_t$, $I_{t, occ}$), we started to observe accuracy gains. We noticed that the combination of $L_{MM}$ and zero forcing produced remarkable performance improvements, possibly a result of \textit{mutual learning} in GRU with the simultaneous flow and occlusion predictions in the availability of context information.
\begin{table}[t]
% \begin{subtable}
\centering
% \vspace{-7mm}
% \caption{Results on Sintel and KITTI train datasets, with C+T datasets}
\caption{Ablation study for Transformation Consistency (TC). H and R are horizontal flips and random rotations (Other notations are the same as Table \ref{tab:abl2}) %\jl{I changed  to Transformation in the title of 3rd column. Please review.}
}
\vspace{-2mm}
\label{tab:abl1}
% \scriptsize
\adjustbox{max width=0.47\textwidth}
{
\begin{tabular}{|@{}l@{}|@{}c@{}|@{}c@{}||@{}c@{}|@{}c@{}||@{}c@{}|@{}c@{}|}
\hline
\multirow{2}{*}{Method (small)} & \multirow{2}{*}{k} &
\multirow{2}{*}{Transformation} & \multicolumn{2}{|@{}c@{}||}{Sintel (train-EPE)}  & \multicolumn{2}{|@{}c@{}|}{KITTI-15 (train)} \\
\cline{4-7}
& & &  Clean & Final & Fl-epe & Fl-all \\
\hline
RAFT (baseline) &-& - & 2.21 & 3.35 & 7.51 & 26.9\\
\hline
\multirow{2}{*}{RAFT + TC} &\multirow{2}{*}{1,2}&  H  & 2.06 & 3.19 & \textbf{6.41} & 22.6\\
& & R & \textbf{2.05} & \textbf{3.15} & 6.50 & \textbf{22.5}\\
\hline
\multirow{2}{*}{RAFT + TC} & 1,2  & \multirow{2}{*}{ R } & \textbf{2.05} & 3.15 & \textbf{6.50} & \textbf{22.5}\\
  & 1,2,3 & & \textbf{2.05} & \textbf{3.14} & 6.69 & 22.6\\
\hline
\end{tabular}
}
\centering
\end{table}
\begin{table}[t]
\centering
\caption{Combination of Transformation Consistency with Occlusion Consistency (Other notations are the same as Table \ref{tab:abl2})}
\label{tab:abl_comb}
\vspace{-2mm}
% \footnotesize
\adjustbox{max width=0.47\textwidth}
{
\begin{tabular}{|l||c|c||c|c|}
\hline
\multirow{2}{*}{Method (small)} & \multicolumn{2}{c||}{Sintel (train-EPE)}  & \multicolumn{2}{|c|}{KITTI-15 (train)} \\
\cline{2-5}
&  Clean & Final & Fl-epe & Fl-all \\
\hline
RAFT (baseline) & 2.21 & 3.35 & 7.51 & 26.9\\
\hline
RAFT + OC & 2.05 & 3.18 & 7.07 & 23.5\\
RAFT + TC &  2.05 & 3.15 & \textbf{6.50} & 22.5\\
RAFT + OC + TC  & \textbf{1.95} & \textbf{3.13} & 6.53 & \textbf{22.1}\\
\hline
% \centering
\end{tabular}
}
\centering
\vspace{-3mm}
\end{table}
\\
\\
\noindent \textbf{Transformation Consistency: }
\label{dis:hori_rot}
We use horizontal flips and random rotations in our transformation consistency strategy, and we evaluate the performance in each type of these transformations\footnote{Some of the transformation methods could potentially improve the performance. Note that rotations ($90^\circ$, $180^\circ$, and $270^\circ$) and horizontal flips guarantee one-to-one correspondences}. As shown in Table \ref{tab:abl1}, the two types of transformations show comparable accuracy gains, although rotation works better empirically in Sintel. Such interesting observations could be attributed to the characteristics of data samples. For example, KITTI image samples are typically dominated by downwards pixel movements in the driving scenes while being quite balanced between rightwards and leftwards movements. This could suggest a strategy to whether apply symmetrical generalization in vertical and horizontal directions. In our supplemental materials, we provide some distribution curves on several datasets.

We also experiment with a range of $k$ values. Within certain $k$ ranges, both Sintel and KITTI samples produce noticeable improvements. It is interesting, however, that Sintel and KITTI empirically demonstrate somewhat different upper bounds for their most suitable $k$ ranges, which could be, again, attributed to the data sample characteristics in flow distributions in vertical and horizontal directions. Systematic analysis may provide more insights into ways of accuracy improvements.
\\
\\
\noindent \textbf{Combining Consistency Strategies: }
In Table~\ref{tab:abl_comb}, both consistency strategies show performance improvements over the baseline model (RAFT-small). And, applying both methods shows better performance. Our conjecture is that the impact of each strategy is enhanced, and generalizability is improved with joint learning. 
\begin{table}[t]
\centering
\caption{Comparisons against the RAFT baseline in accuracy, model size, and inference time on KITTI after 24 GRU iterations.}
\label{tab:size_speed}
\vspace{-2mm}
% \footnotesize
\adjustbox{max width=0.47\textwidth}
{
\begin{tabular}{|l||c|c||c|c|}
\hline
\multirow{2}{*}{Model} & \multicolumn{2}{c||}{KITTI} &$\#$ of  & \multirow{2}{*}{Inference Time} \\
& Fl-epe & Fl-all & Parameters & \\
% Model & $\#$ of Parameters & Speed\\
\hline
\hline
RAFT (small) &7.51 & 26.9 & 990,162 & 99.03 ms \\
RAFT + OCTC (small) & 6.53 & 22.1 & 997,043 & 101.53 ms \\
\hline
RAFT & 5.04 & 17.4 & 5,257,365 & 140.18 ms \\
RAFT + OCTC & 4.72 & 16.3 & 5,263,803 & 143.21 ms \\
\hline
% \centering
\end{tabular}
}
\centering
\vspace{-3mm}
\end{table}
\\
\\
%\noindent \textbf{\js{Restore Transformation}: }
\noindent \textbf{Transformation Restoration: }
We considered inverting not only the displacement quantities but also the signs and axes when restoring coordinates back from transformation. For example, in restoring the $90^\circ$ rotation, we computed the inverse of the pixel location and changed the signs and flow vector axes.
\vspace{2mm}
\\
\noindent \textbf{Model Size and Speed: }
%\subsection{Model Size and Memory}
We measure the average inference times with KITTI dataset using Nvidia V100DX-8C GPU. Our models significantly outperform the baseline RAFT at only minimal model overhead as detailed in Table~\ref{tab:size_speed}. To support transformation consistency, there is no model size increase. Occlusion consistency entails minor model size increases by only 0.12\% and 0.69\% on large and small models, respectively, for mask derivation, which also has a minimal impact on inference time. Besides, during training, our model computes the baseline and transformation outputs sequentially without needing extra memory.  
\vspace{2mm}
\\
\noindent \textbf{Limitations:}
Our algorithm could be further improved to work for very large areas of occlusions. Besides, we currently use only self-supervised learning in our occlusion training with sample pairs created from individual images ($I_t$, $I_{t,occ}$). Furthermore, we speculate that it could be challenging to predict accurate optical flows in certain low-frequency regions, where boundaries may be hidden due to occlusion. This problem could be investigated using an occlusion generating network with labeled data.

Another area of further research for improvement could be an analysis on the frame rate. Beyond our methods of consistency, zero forcing, and frame hopping, aspects such as temporal consistency could be investigated.

\section{Conclusion}
\label{sec:con}

% \fp{this parts should be revised thoroughly...}

In this paper, we have introduced novel and effective consistency learning strategies, promoting occlusion consistency and transformation consistency, for optical flow estimation. We further introduce enhancements, zero forcing as a special case of occlusion consistency and frame hopping as a generalization to transformation consistency, to our overall consistency learning framework. Applying these methods jointly, we demonstrate empirical outperformance over the baselines. Specifically, our method sets the new state-of-the-art performance and has ranked top in the KITTI-2015 scene flow non-stereo leaderboards. We intend to adapt our framework to wider tasks in our future study.

\newpage

%%%%%%%%% REFERENCES
{\small
\bibliographystyle{ieee_fullname}
\bibliography{egbib}
}

\clearpage
\newpage

\section{Appendix}
% \begin{appendices}

\subsection{Datasets:}
In our experiments, we have utilized the FlyingChairs (C) \cite{dosovitskiy2015flownet},
FlyingThings3D (T) \cite{mayer2016large},
Sintel (S) \cite{butler2012naturalistic}, and
KITTI (K) \cite{geiger2013vision, menze2015object} datasets which are the most popular datasets in the optical flow estimation problem.
FlyingChairs \cite{dosovitskiy2015flownet} consist of 22,872 image pairs and the corresponding ground truths.
It is composed of individual pairs, so we cannot constitute  additional image pairs corresponding to $k > 1$.
FlyingThings3D \cite{mayer2016large} consists of a training dataset of 21,818 images and a test dataset of 4,248 images.
The images of FlyingThings3D consist of more than two consecutive frames, which have both the forward optical flow (20,151 pairs) and the backward optical flow (20,151 pairs) ground truth.
In addition, this and Sintel datasets are categorized into clean pass and final pass, and the final pass is applied a post-processing such as fog impact, motion blur, and so on. 
Therefore, the number of pairs in the training set of FlyingThings3D dataset increases to 80,604.
Sintel \cite{butler2012naturalistic} consists of a training dataset of 1,064 images and a test dataset of 564 images.
Sintel is also composed of more than two consecutive frames, and as mentioned above, it is composed of a clean pass and a final pass.
KITTI \cite{geiger2013vision, menze2015object} consists of a training dataset of 400 images and a test dataset of 400 images.
HD1K \cite{kondermann2016hci} consists of 1,083 images.
These are composed of individual pairs same as FlyingChairs, so there are 200 pairs in both training and test datasets.

\begin{table}[h]
\begin{center}
\vspace{-3mm}
\caption{We perform hyperparameter search over a grid of $\lambda_{1}$ $\in$ \{1.0, 0.1, 0.01, 0.001\} in Eq.10.
% The parameters are set to $\alpha_{2}$ = 0.0.
We trained the model with the Flyingchairs (C) and Flyingthings (T) datasets and tested the model on the training dataset of the Sintel (S) and KITTI (T).}
\label{tab:abl3}
% \vspace{-3mm}
\footnotesize
{

\begin{tabular}{|c|c||c|c||c|c|}
% \begin{tabular}{|@{}l@{}|l@{}||c|@{}c@{}||c|c|}
\hline
Method & \multirow{2}{*}{\large{$\lambda_{2}$}} & \multicolumn{2}{|@{}c@{}||}{Sintel (train-EPE)}  & \multicolumn{2}{|@{}c@{}|}{KITTI-15 (train)} \\
\cline{3-6}
(small) & &  Clean & Final & F1-epe & F1-all \\
\hline
% \hline
% \multicolumn{6}{|c|}{small}\\
% \hline
RAFT  & - & 2.21 & 3.35 & 7.51 & 26.9\\
\hline
\multirow{4}{*}{RAFT + OC} & 1.0 & 2.48 & 3.60 & 8.57 & 27.6\\
 & 0.1 & \textbf{2.05} & \textbf{3.18} & \textbf{7.07} & \textbf{23.5}\\
 & 0.01  & 2.19 & 3.24 & 7.41 & 23.6 \\
 & 0.001 & 2.24 & 3.26 & 7.52 & 25.0\\
 
\hline
\end{tabular}
}
\end{center}
% \vspace{-5mm}
\end{table}

\begin{table}[h]
\begin{center}
\vspace{-3mm}
\caption{We perform hyperparameter search over a grid of $\lambda_{2}$ $\in$ \{1.0, 0.1, 0.01, 0.001\} in Eq.10.
The parameters are set to Transformation = R, $\epsilon$ = 25.0, and k = 1,2.
We trained the model with the Flyingchairs (C) and Flyingthings (T) datasets and tested the model on the training dataset of the Sintel (S) and KITTI (T).
}
% \vspace{-3mm}
\label{tab:abl2}
\footnotesize
{
\begin{tabular}{|c|c||c|c||c|c|}
% \begin{tabular}{|@{}l@{}|l@{}||c|@{}c@{}||c|c|}
\hline
% Method & Labeled & Network & \multicolumn{2}{|c|}{mAP (\%)}\\
% \vspace{2mm}
Method & \multirow{2}{*}{\large{$\lambda{1}$}} & \multicolumn{2}{|@{}c@{}||}{Sintel (train-EPE)}  & \multicolumn{2}{|@{}c@{}|}{KITTI-15 (train)} \\
\cline{3-6}
(small)& &  Clean & Final & F1-epe & F1-all \\
\hline
% \hline
% \multicolumn{6}{|c|}{small}\\
% \hline
RAFT & - & 2.21 & 3.35 & 7.51 & 26.9\\
\hline
\multirow{4}{*}{RAFT + TC} & 1.0 & 3.05 & 3.87 & 13.41 & 34.7\\
 & 0.1 & 2.06 & 3.23 & 7.16 & 23.3 \\
 & 0.01  & \textbf{2.05} & \textbf{3.15} & 6.50 & \textbf{22.5} \\
 & 0.001 & \textbf{2.05} & 3.20 & \textbf{6.47} & 22.7\\

\hline
\end{tabular}
}
\end{center}
\vspace{-3mm}
\end{table}

\begin{table}[h]
\begin{center}
\vspace{-3mm}
% \caption{Results on Sintel and KITTI train datasets, with C+T datasets}
\caption{We perform hyperparameter search over a grid of epsilon $\epsilon$ $\in$ \{$3^{2}$, $5^{2}$, $7^{2}$, $\infty$\} in Eq.8 under Transformation Consistency setting.
The parameters in Transformation Consistency are set to $\lambda_{2}$ = 0.01, Transformation = R, and k = 1,2.
We trained the model with the Flyingchairs (C) and Flyingthings (T) datasets and tested the model on the training dataset of the Sintel (S) and KITTI (T). 
}
% \vspace{-3mm}
\label{tab:abl}
\footnotesize
{
\begin{tabular}{|c|c||c|c||c|c|}
% \begin{center}
% \begin{tabular}{|@{}l@{}|l@{}||c|@{}c@{}||c|c|}
\hline
% Method & Labeled & Network & \multicolumn{2}{|c|}{mAP (\%)}\\
% \vspace{2mm}
Method & \multirow{2}{*}{\large{$\epsilon$}} & \multicolumn{2}{|@{}c@{}||}{Sintel (train-EPE)}  & \multicolumn{2}{|@{}c@{}|}{KITTI-15 (train)} \\
\cline{3-6}
(small)& &  Clean & Final & F1-epe & F1-all \\
\hline
% \hline
% \multicolumn{6}{|c|}{small}\\
% \hline
RAFT & - & 2.21 & 3.35 & 7.51 & 26.9\\
\hline
\multirow{4}{*}{RAFT + TC} & $3^{2}$ & 2.09 & 3.19 & \textbf{6.46} & \textbf{22.5}\\
 & $5^{2}$ & 2.05 & \textbf{3.15} & 6.50 & \textbf{22.5} \\
 & $7^{2}$ & \textbf{2.04} & 3.16 & 6.63 & 22.6\\
 & $\infty$ & 2.09 & 3.18 & 6.91 & 22.9\\
\hline

\end{tabular}
}
\end{center}
\vspace{-3mm}
\end{table}

\subsection{Implementation Details: }
The codes used for our experiments are based on Pytorch, and we have used the official code\footnote{\url{https://github.com/princeton-vl/RAFT}} for RAFT \cite{teed2020raft}.
Our method introduces three additional hyper parameters, namely, ($\lambda_{1}$, $\lambda_{2}$) of Eq.10 and $\epsilon$ of Eq.8.
We performed a grid search over the values in \{1.0, 0.1, 0.01, 0.001\} for each $\lambda$ in Eq.10 and in \{$3^{2}$, $5^{2}$, $7^{2}$, $\infty$\} for $\epsilon$ value in Eq.8.
In table~\ref{tab:abl3}, our model with occlusion consistency shows best performance at $\lambda_{1}$ = 0.1. For transformation consistency, our model shows superior scores in most evaluations at $\lambda_{2}$ = 0.01. In case of the $\epsilon$, our transformation consistency loss has shown good performance in Sintel dataset with ($5^2$ and $7^2$ for $\epsilon$) and in KITTI dataset with ($3^2$ and $5^2$ for $\epsilon$). Therefore, we set the parameters to be [($\lambda_{1}$, $\lambda_{2}$) = (0.1, 0.01), $\epsilon$ = $5^{2}$].

\onecolumn
\newpage

\subsection{Dataset Characterization with Displacement Distributions}

Fig.~\ref{exp:data_distribution} below shows cumulative density functions (CDFs) of the ground truth displacements for four popular optical flow datasets. In each plot, we accumulate displacement values symmetrically from $-100$ to $100$ for individual dimensions of $(u,v)$, corresponding to the $X$ and $Y$ axes, excluding larger displacements as outliers.
For the FlyingChair dataset, the figure shows that most of the samples are near zero with a relatively small variance.
The FlyingThings3D dataset, instead, shows a larger variance than FlyingChair and Sintel.
In addition, KITTI appears to have a larger variance than the other datasets, possibly due in part to its smaller frame rates used in this dataset.
Another interesting observation from the figures is that, unlike other datasets, KITTI demonstrates significant asymmetry in the flow distribution on the $Y$ axis, as the images are probably dominated by downward movements in the images captured with frontal views of the vehicles.

\begin{figure*}[h]
\begin{center}$
\centering
\begin{tabular}{c c c c }

% \textbf{FlyingChairs} &  & \textbf{FlyingThings3D} & \\
\multicolumn{2}{c}{\textbf{FlyingChairs}} & \multicolumn{2}{c}{\textbf{FlyingThings3D}}\\
% RAFT-S$^4$L RAFT-S4L
\includegraphics[width=4.2cm]{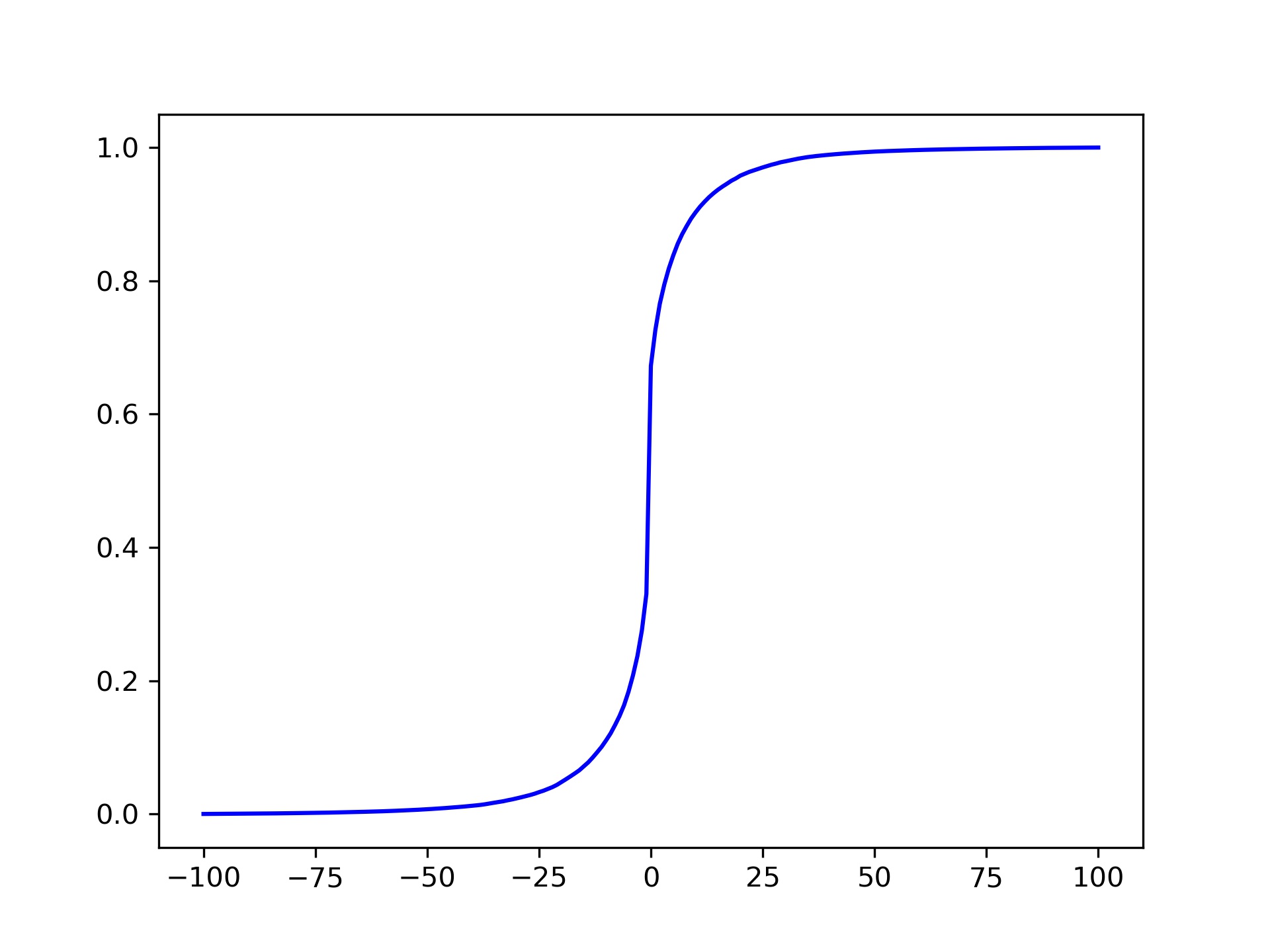} & \hspace{-10mm}
\includegraphics[width=4.2cm]{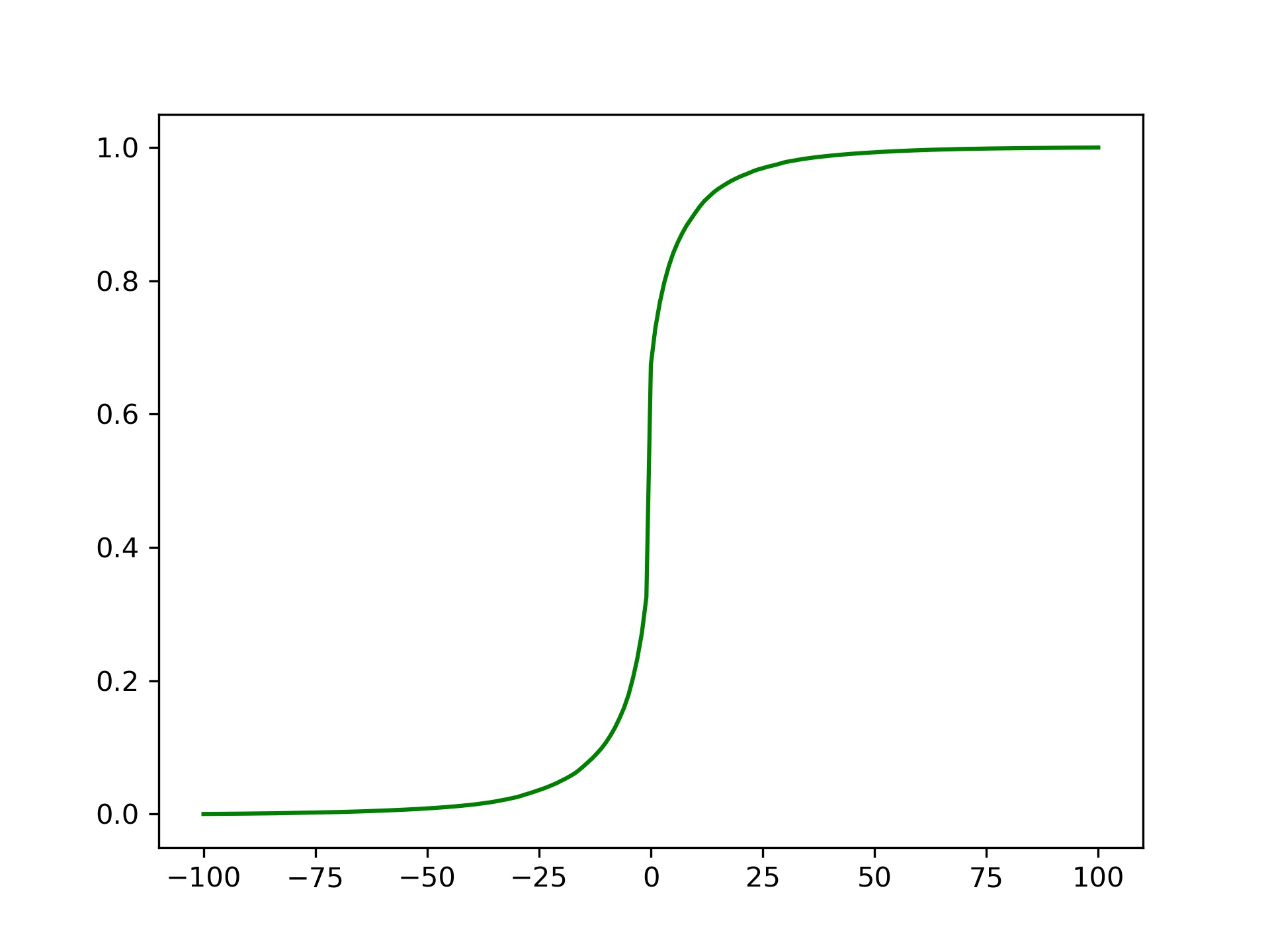} &
\includegraphics[width=4.2cm]{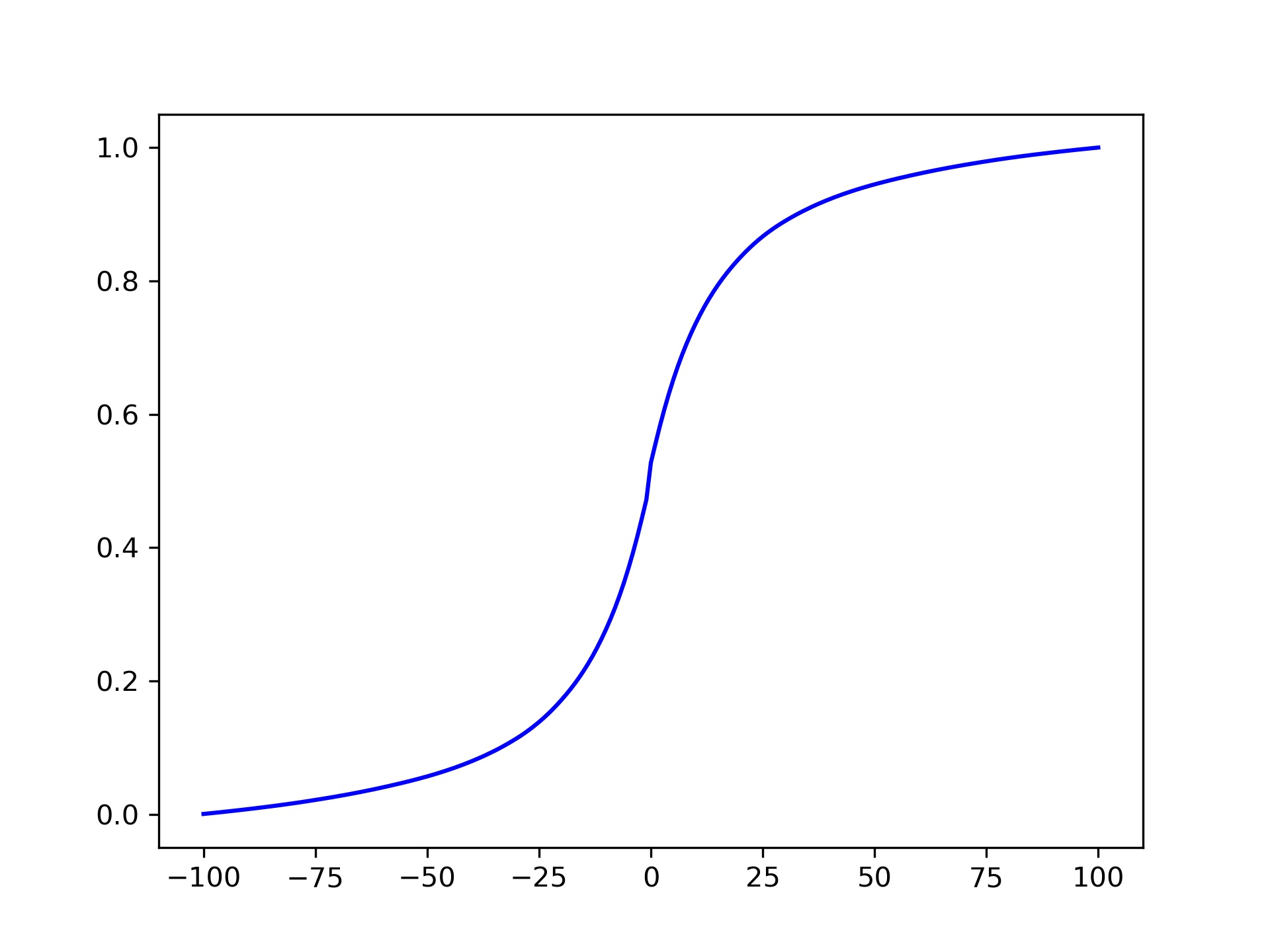} & \hspace{-10mm}
\includegraphics[width=4.2cm]{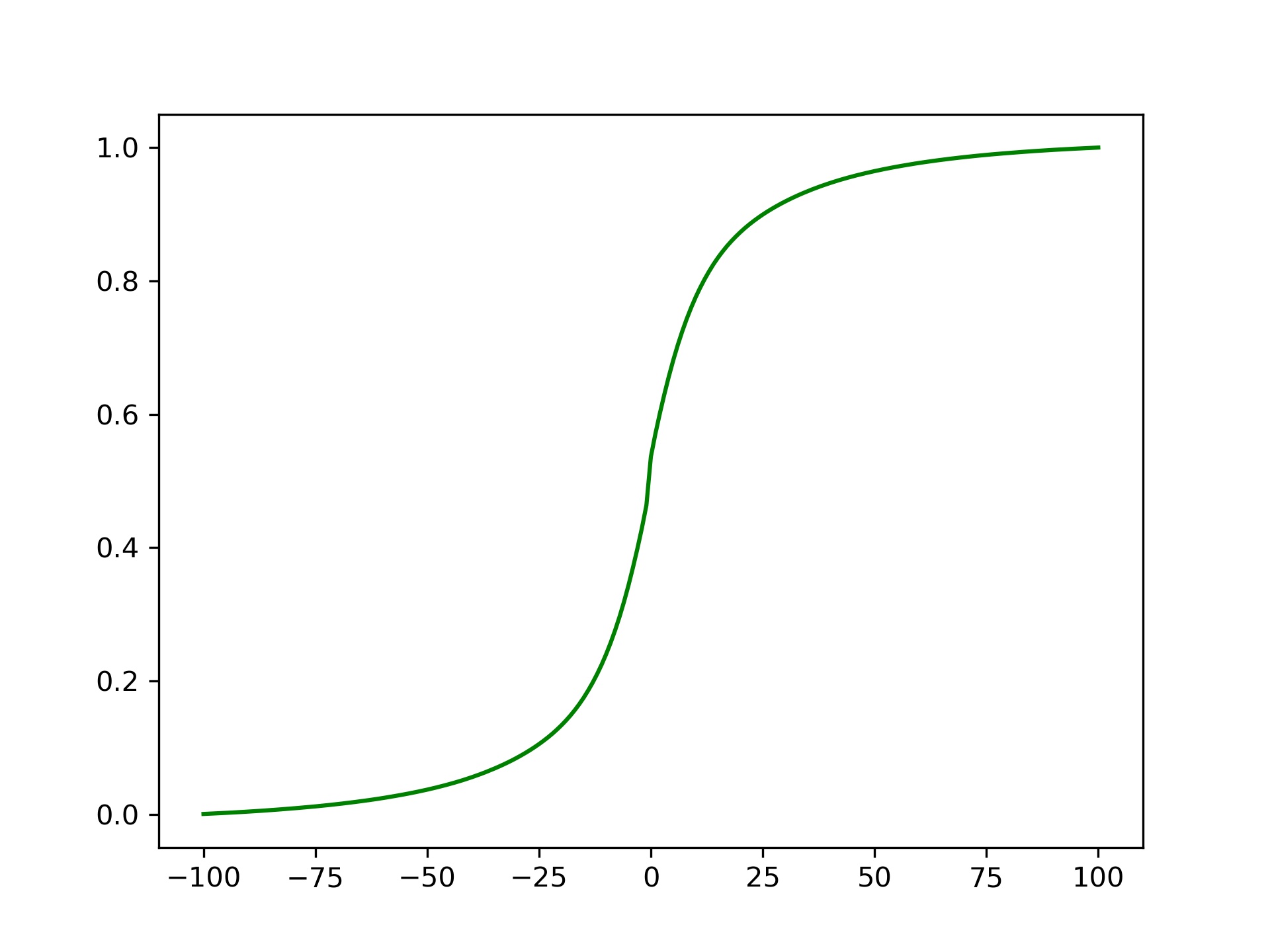} \\
% \vspace{-2mm}
\text{u} &\hspace{-10mm} \text{v} & \text{u} &\hspace{-10mm} \text{v} \\
% \vspace{10mm}
\\
\multicolumn{2}{c}{\textbf{Sintel}} & \multicolumn{2}{c}{\textbf{KITTI}}\\
% RAFT-S$^4$L RAFT-S4L
\includegraphics[width=4.2cm]{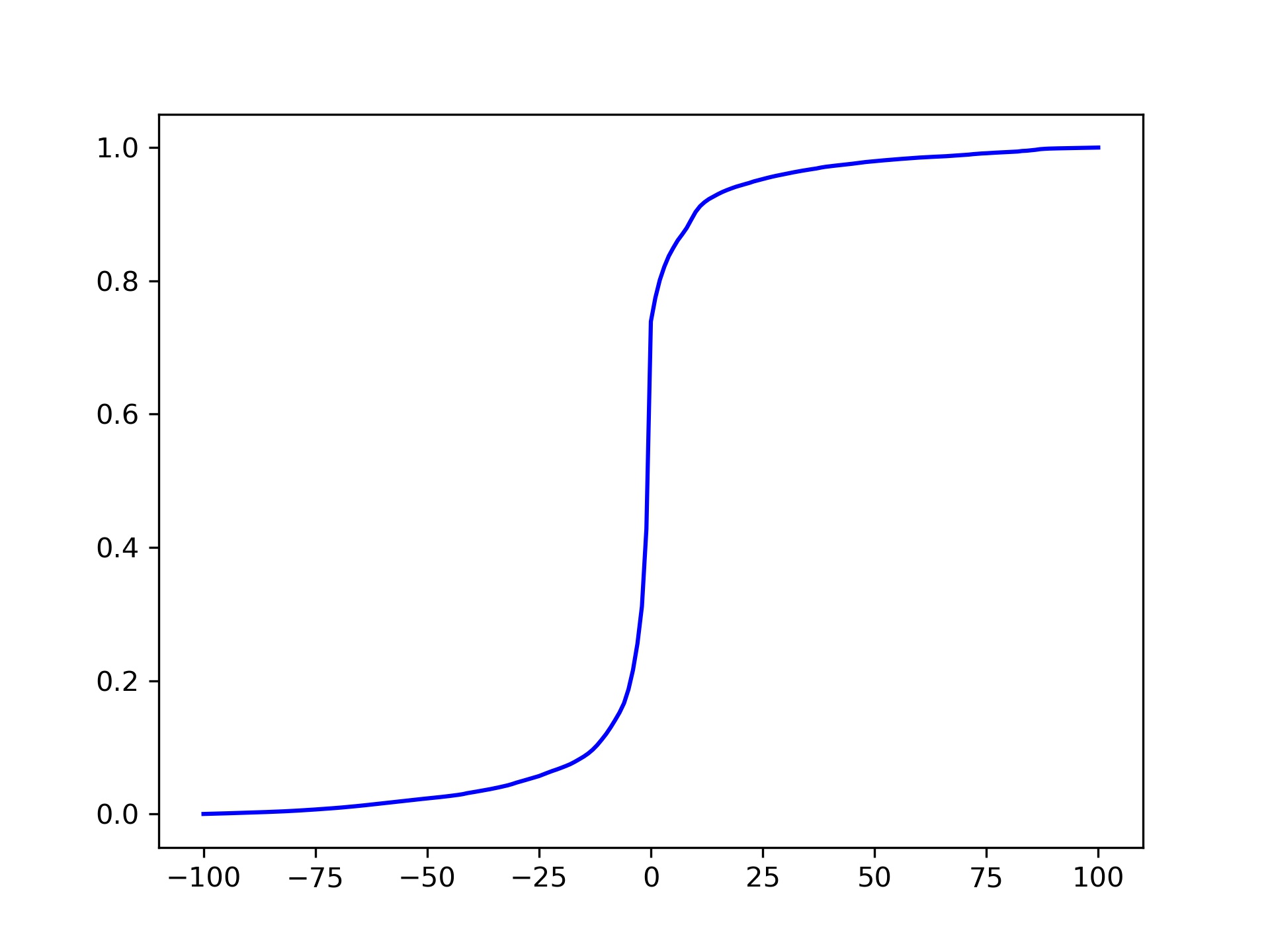} & \hspace{-10mm}
\includegraphics[width=4.2cm]{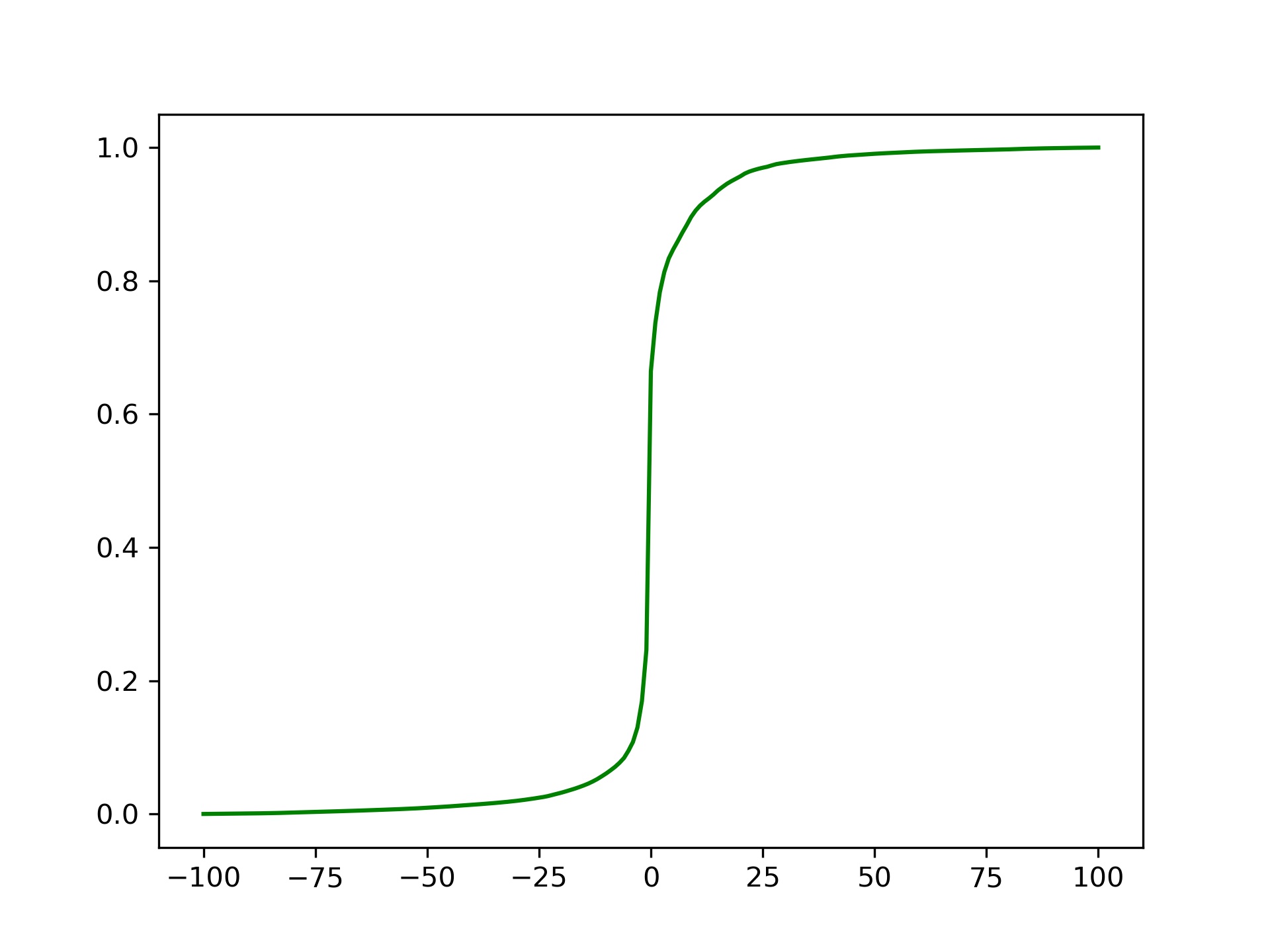} &
\includegraphics[width=4.2cm]{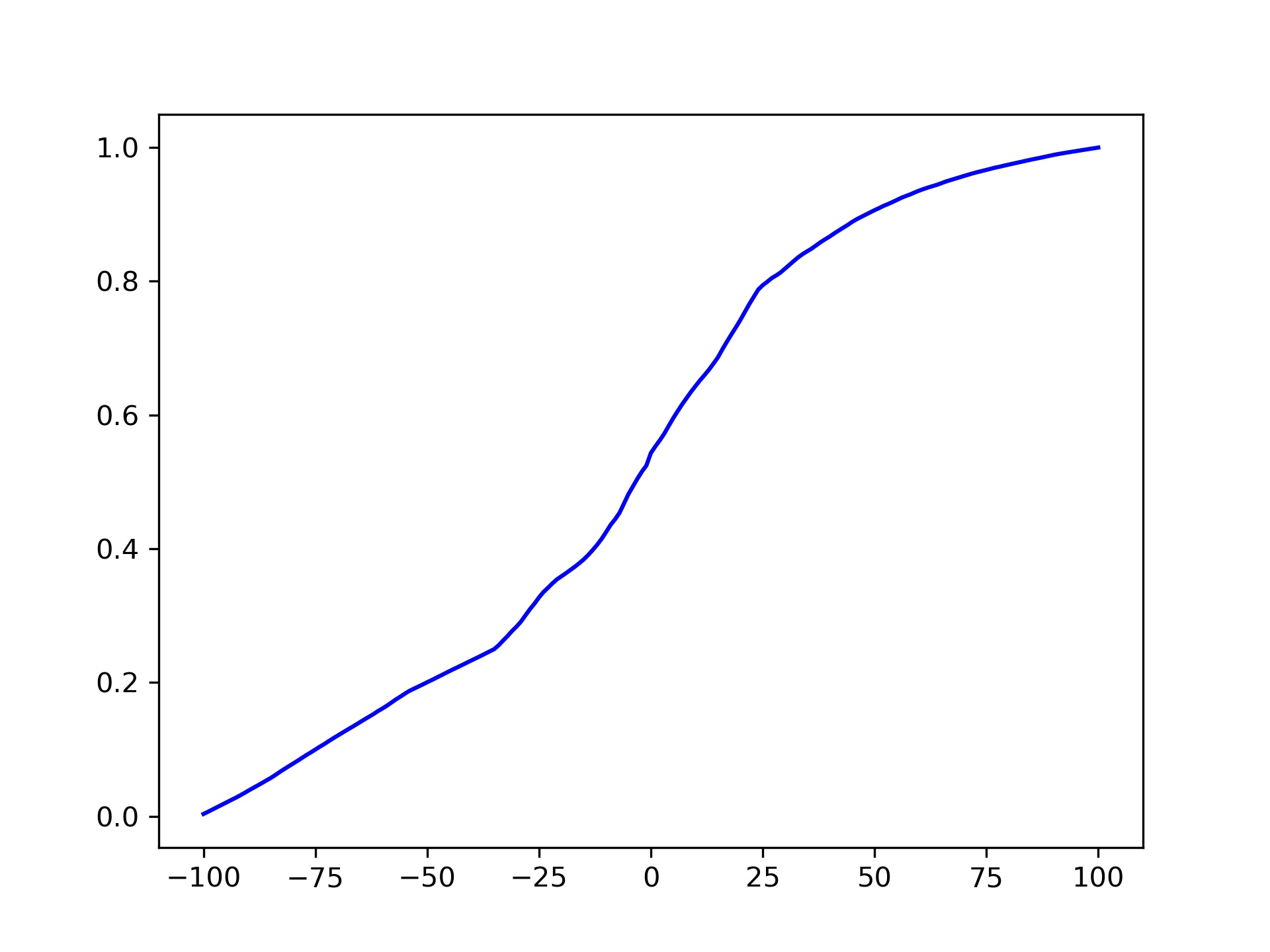} & \hspace{-10mm}
\includegraphics[width=4.2cm]{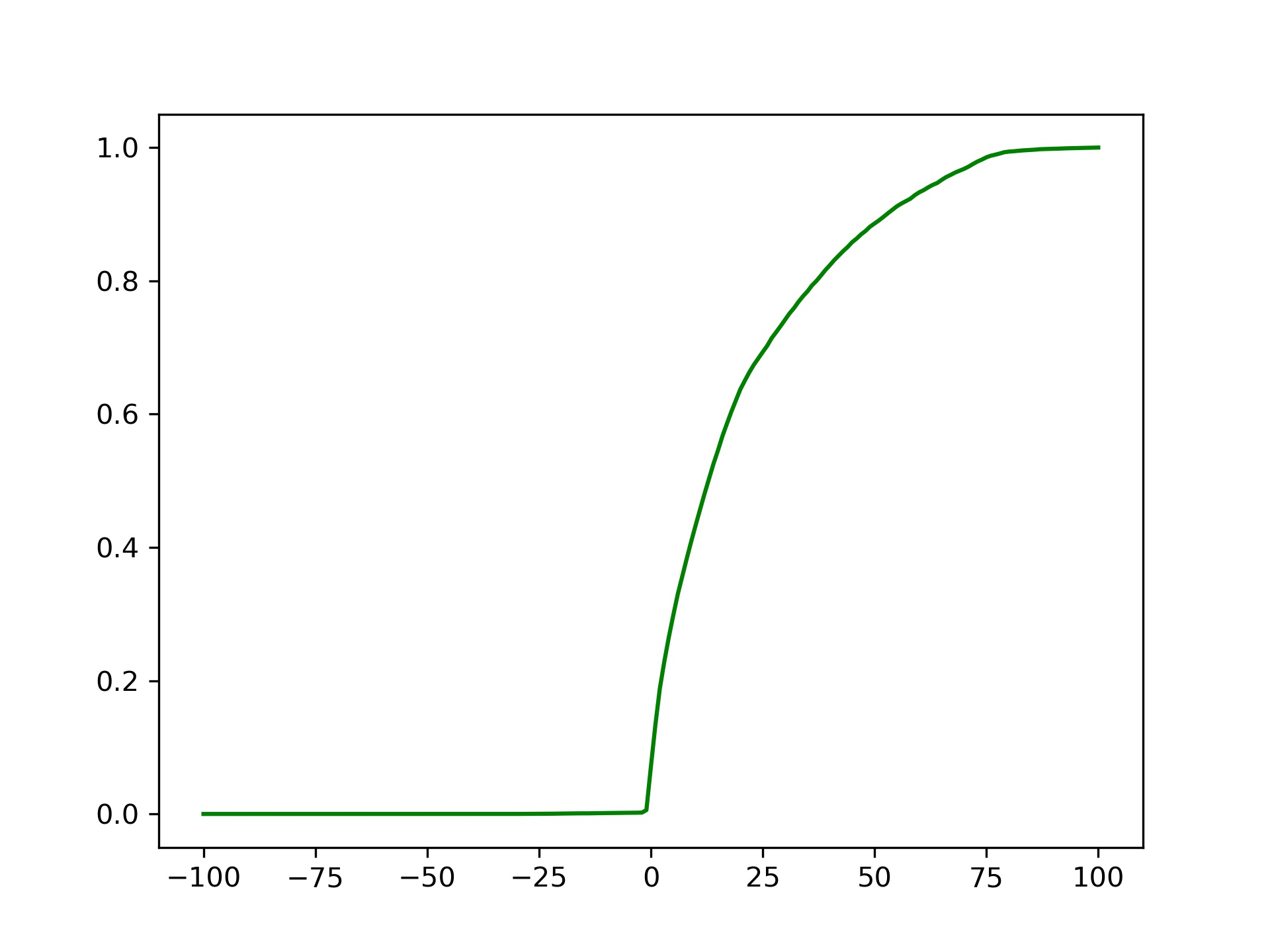} \\
\text{u} &\hspace{-10mm} \text{v} & \text{u} &\hspace{-10mm} \text{v} \\

% \vspace{-2}
\end{tabular}$
\end{center}
\vspace{-2mm}
\caption{Cumulative Density Functions (CDFs) of displacements in popular datasets. In each pair of sub-figures, the left and rigtht sub-figures show displacements on the $X$ ($u$) and $Y$ ($v$) axes, respectively.
}
\label{exp:data_distribution}
\vspace{-1mm}
\end{figure*}

% \onecolumn
% \newpage
\newpage

\subsection{Performance Comparisons with More Examples}

In the figure on the next page, we provide performance comparison with additional examples in a range of various EPEs from low to high.

% \newpage

\begin{figure*}[h]
\begin{center}$
\centering
\begin{tabular}{c}

\includegraphics[width=\linewidth]{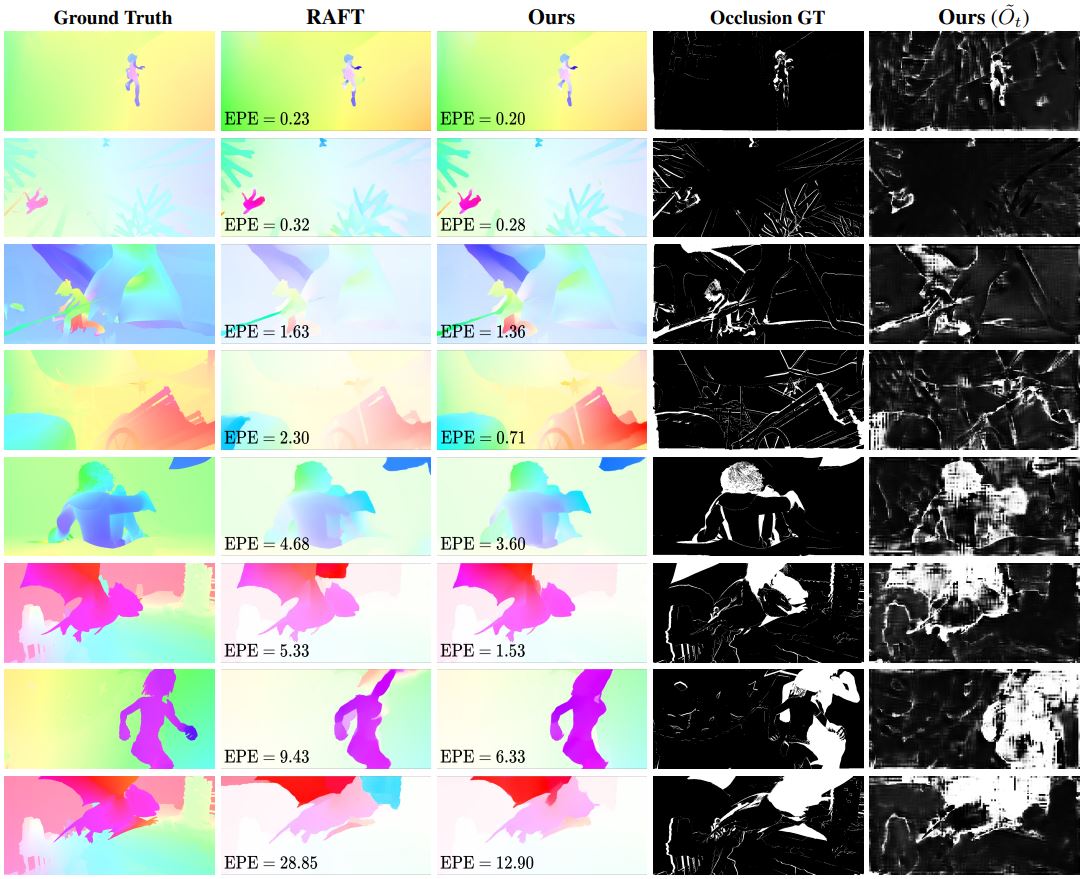}

\vspace{-4mm}
\end{tabular}$
\end{center}
% \vspace{-2mm}
\caption{More performance comparisons between the baseline (RAFT) and Ours (RAFT-OCTC) on Sintel train samples (trained with C+T).
% \vspace{-3mm}
}
\label{exp:example_image}
% \vspace{-1mm}
\end{figure*}
% \end{appendices}

\end{document}